\documentclass[12pt]{article}
\usepackage[utf8]{inputenc}
\usepackage{indentfirst}
\usepackage[demo]{graphicx}
\usepackage{caption}
\usepackage{booktabs}
\usepackage{subcaption}
\usepackage{array}
\usepackage{appendix}
\title{Final Year Project Report}
\author{qian.zheng14 }
\date{January 2019}

\usepackage{hyperref}
\hypersetup{
    colorlinks,
    citecolor=black,
    filecolor=black,
    linkcolor=black,
    urlcolor=black
}

\begin{document}

\begin{titlepage}

\newcommand{\HRule}{\rule{\linewidth}{0.5mm}} 
\setlength{\topmargin}{0in}
\center 

 \begin{minipage}{0.4\textwidth}
\begin{flushleft} \large
\hspace*{-0.5cm}
\end{flushleft}
\end{minipage}
~
\begin{minipage}{0.5\textwidth}
\begin{flushright} \large
\hspace*{2cm}
\end{flushright}
\end{minipage}\\[1cm]

\textsc{\LARGE Imperial College of London}\\[1.5cm] 
\textsc{\Large Department of Computing}\\[0.5cm] 


\HRule \\[0.4cm]
{ \huge \bfseries Emotion recognition with 4k resolution database}\\[0.4cm] 
\HRule \\[1cm]
 

\begin{minipage}{0.4\textwidth}
\begin{flushleft} \large
\emph{Author:}\\
Qian \textsc{Zheng} \\ 
\end{flushleft}
\end{minipage}
~
\begin{minipage}{0.5\textwidth}
\begin{flushright} \large
\emph{Imperial College Supervisor:} \\
Dimitrios \textsc{Kollias} \\[0.5cm] 
\end{flushright}
\end{minipage}\\[1cm]


{\large \today}\\[0.5cm] 

\vfill 

\end{titlepage}
\begin{abstract}
Classifying the human emotion through facial expressions is a big topic in both the Computer Vision and Deep learning fields. Human emotion can be classified as one of the basic emotion types like being angry, happy or  dimensional emotion with valence and arousal values.  There are a lot of related challenges in this topic, one of the  most famous challenges is called the 'Affect-in-the-wild Challenge' (Aff-Wild Challenge). It is the first challenge on the estimation of valence and arousal in-the-wild. This project is an extension of the Aff-wild Challenge. Aff-wild database was created using images with a mean resolution of $607 \times 359$, me and Dimitrios sought to find out the performance of the model that is trained on a database that contains 4K resolution in-the-wild images. Since there is no existing database to satisfy the requirement, I built this database from scratch with help from Dimitrios, and trained neural network models with different hyperparameters on this database. I used network models like VGG16, AlexNet, ResNet and also some pretrained models like ImageNet VGG. I compared the results of the different network models alongside the results from the Aff-wild database to exploit the optimal model for my database.

\end{abstract}
\newpage
\newpage
\renewcommand{\abstractname}{Acknowledgements}
\begin{abstract}
I want to express my gratitude to my supervisor, Dimitrios Kollias. He has guide me a lot through this project, and we keep contacting to make sure I am always on the right track of this project. I would also like to thank Dimitrios's teammates, as they helped me with annotating the whole database in a very professional manner. 
\end{abstract}
\newpage
\tableofcontents
\section{Introduction}

Emotion is a representation of a human's mental state, and it can help people with social interaction. People are used to recognising another person's emotions and adjusting their behaviour based on that judgement. However, compared to trying to understand their emotion through intonation or body language, people are more comfortable with analysing facial expressions as it's much easier and more practical. Automatically analysing facial expressions has become a state of art technique in computer vision, and this project attempts to extend the functionality and improve the performance of this technique.

Most of the research in facial analysis is done using datasets of posed behaviour which are captured under controlled recording conditions. \cite{controlled} However, recently researchers started to focus on using in-the-wild databases which means the images are taken spontaneously. It's obvious that facial expressions of people under controlled environments can be different from the expressions under real-world conditions with the same emotions. So instead of using posed images as input data, I decided to use in-the-wild images.

There have been projects using in-the-wild images to construct models\cite{Aff-wild}, but like other research in this area, most of the images used for facial analysis have low or normal resolution(800x600). Normally, the resolution of the images can't be too small as it could lose important information. The resolution also shouldn't be too big either, as the dimensions and size of neural networks can grow significantly as the resolution does. However, high resolution images contain more information than normal images, if a model can be trained using a 4k resolution database, the model may outperform other existed models. The purpose of this project, is to research on whether the latest models are able to handle 4k images as input, and will it be able to classify the emotion accurately.

\newpage
\section{background}
\subsection{Human emotions and emotion recognition}
Emotion is a mental state related with a human's behaviour, thoughts and feelings.\cite{emotion} It was widely accepted that human emotion can be classified into six basic emotions: surprise, fear, disgust, anger, happiness and sadness. \cite{Emotions} 

In Scherer's components processing model of emotion, a person's emotions requires a period of time in order for them to fully processes the five different components synchronously and coordinately. These components are \cite{Emotions} :
\begin{itemize}
  \item Cognitive appraisal: provides an evaluation of events and objects.
  \item Bodily symptoms: the physiological component of emotional experience.
  \item Action tendencies: a motivational component for the preparation and direction of motor responses.
  \item Expression: facial and vocal expression almost always accompanies an emotional state to communicate reaction and intention of actions.
  \item Feelings: the subjective experience of emotional state once it has occurred.
\end{itemize}

Researchers have found that expression is something that always accompanies an emotional state. Therefore, in order to recognise emotions, firstly one needs to analyse their expression. People are used to predicting someone else's emotions through their verbal and facial expressions. Someone who is angry has a different facial expressions compared to people who are sad. Furthermore, positive emotions always lead to a larger volume of speech than negative emotions do. These expressions can be easily captured by humans during social activities and events. Therefore there is a strong belief that machines are capable of recognising and analysing facial expressions for these emotions as well.

Emotion recognition has already become a wildly used technique, it has helped with different tasks in different industries. One of the biggest areas it is used in is marketing research. Instead of collectively reviewing products by only using surveys or questionnaires, Disney has been using emotion recognition techniques to obtain information about the feelings of audiences whilst they're watching their films.\cite{Disney} Their software is able to simultaneously capture the facial expressions of more than 3000 members of an audience. This is very much beyond the capability of what a human being could do. This software means that Disney are able to have a better understanding of the emotional impact of their films, hence reflecting upon the quality of the movie. Adjusting the content of the film according to this emotional impact. 

This technique has also been applied on candidate interviews. Lots of companies are using HireVue AI-powered technology to capture a candidate's body language and mood while they answer interview questions. The company can then find out if the candidate is suitable for the role.

The Gaming industry is also using this technology as well. Companies can capture game testers emotions whilst they are testing the video game. According to the emotional feedback of those players, the company can better decide how they can improve content of the game. For example, the level of anxiety and fear is often important for a horror game.  

I believe there are many more possibilities in the uses of emotion recognition \cite{tag,park,deep}. Thus, extending the capability of this technology is the main contribution of this paper. 

\subsection{Facial expression analysis}
Facial expression is one of the main information channels that can be used to understand the emotions of different people and this has drawn a lot of attention in the last two decades. Advanced computer vision techniques \cite{eccv,ijcv} have played a critical role to analysing the facial expression and the most common way is to capture it, annotate it and finally classify it. 

Pyramid Histogram of Oriented Gradients (PHOG) \cite{PHOG} is a descriptor that captures information about the gradient orientations in localised areas of an image. There are also other descriptors like Action units and boosted LBP descriptors. \cite{LBP} Descriptors are used to extract the features from images and those features will be fed into classifiers in order to classify the emotion.

Facial images are annotated using discrete or dimensional emotion labels such as sadness or anger. The classifier will be trained using input data and it will learn how to classify unseen facial images as one of the annotated labels. 

One of the most famous classifiers is the Support Vector Machine (SVM). It's a  supervised learning model with learning algorithms that analyse data for classification\cite{SVM1}. It classifies the data by constructing a set of support vectors, these vectors outline the hyperplanes among the training vectors that maximise the distance between each class given to the SVM. 

Another type of classifier is a deep learning neural network. Neural networks are a type of computing system that is inspired by biological neural networks within the brain of an animal \cite{NN}. An ideal neural network is considered as a universal function. Yet, the network will need to spend time on learning to solve the problem by giving examples of the solution or figure it out itself by taking rewards and punishment. A much more thorough explanation of neural networks will be discussed later in this chapter. 

In the next few subsections, I will discuss how to annotate the emotion of a face in a given image; the existing databases that are available for these images and annotations and finally what types of neural network (classifier) can be used for this purpose.

\subsection{Emotion annotation}
In general, people classify emotions into six basic emotions, we call these discrete emotions. However, researchers believe that these basic emotions can't express all the facial expressions of a human. To deliver a more accurate recognition method, instead of using basic emotions, scientists have developed several different ways to annotate the facial expression of different emotions.

\subsubsection{Action units}
Facial action coding system (FACS) is  a system based on the facial muscle changes to express the human emotion.\cite{FACS} The movements of the muscle are encoded as units called action units(AUs). Cheek Raiser is encoded as 6 and Blink is encoded as 45. In total,there are 46 fundamental units. A group of actions units can be seen when people producing certain emotion. For example units 4,15 can be seen when the emotion is Sadness.4,7,24 can be seen when the emotion is angry\cite{AU}. Based on the FACS and AUs, one popular way to annotate the emotion for image is to detect the Action unit, once we have these unit, we can classify the emotion based on them.

However,automatic recognition of action units is a difficult problem. That's because AUs have no quantitative definitions and as noted can appear in complex combinations.\cite{AU for facial expression}To solve those problems, one suitable approach is to pose AU detection as a binary- or multi-class classification problem using different features and classifier (eg SVM)\cite{AU approach ML}.

Moreover, detecting AU from videos also face some challenges as well. For example how to cope with large variability of action units across the video.Fortunately, the dynamic Bayesian network can address this issue by  modelling the dynamic AUs as a transition in a  state space. \cite{Dynamic AU}

\subsubsection{valence arousal model}
Another method for emotion annotation is using a dimensional emotion representation. \cite{dimentional} . Researchers have noticed that when people describing their emotion, it's difficult to suggest that they are feeling one specific emotion isolated. Instead, they prefer to recognise their emotions as a overlapping experience\cite{Varieties of anger}. One example is most of people describing one specific positive emotion along with some other positive emotion.\cite{General and specific factors}. However the ambiguous and overlapping between basic emotions can be addressed using dimensional model because the dimensional model can regard all those experience as a continuous states.

The most famous dimensional emotion representation is valence-arousal model. \cite{circumplex model} This model suggests that emotions are distributed in a 2D circular space, containing arousal and valence dimensions. Arousal represents the vertical axis and valence represents the horizontal axis, whereas the arousal can varies from active to passive and valence varies from positive to negative. While the centre of the circle represents a neutral valence and a medium level of arousal. In this model, emotional states can be represented at any point in this space with specific vertical and horizontal axis.\cite{A comparison of dimensional models}.

\begin{figure}[h]
	\centering
    \includegraphics[width={1\textwidth},height={0.6\textwidth}]{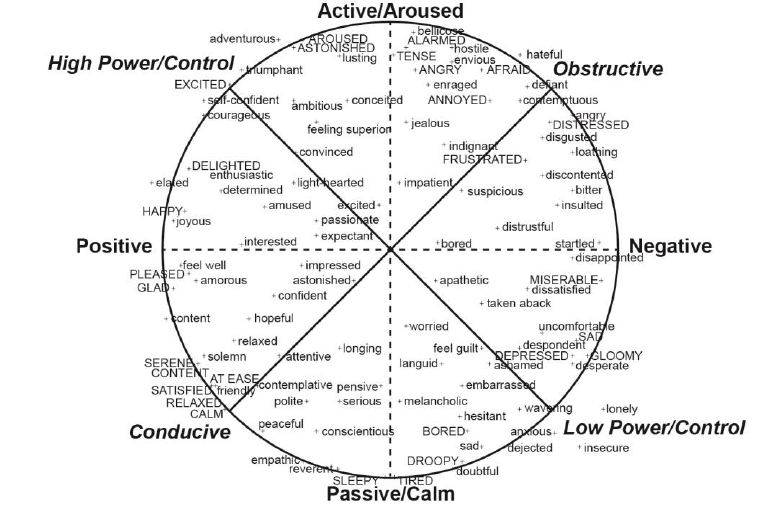}
    \caption{Valence-arousal 2D wheel}
    \label{fig:clean_curve}
\end{figure}

I chose to use Valence-arousal model to annotate the dataset. That's becuase the dataset we constructed are frames of videos. The facial expression varies along with the timestamp which means the valence and arousal values are continuous. Valence-Arousal model is designed for continuous emotion states and I think it's perfect for my database.
\subsubsection{Tools for annotation}
To annotate the value for input data, I will need to use tools. The purpose of this process is to label the  data, which means give each frame a valence value and a arousal value. I intended to use the dataset with more than 1 million frames so it's a time consuming task to annotate every frames manually. I need to use a tool to annotate the each frames time-continuously.

There are some existed tools to do this job such as Feeltrace\cite{Feeltrace} and Gtrace\cite{Gtrace}. But Dimitrios Kollias has built a similar tool with his team and he allows me to use this tool, so I decided to use it follow his guideline. The annotation process is as follow\cite{Aff-wild}:

\begin{itemize}
  \item logs into application and select the joystick
  \item all the imported videos would appear and user should select the one to annotate
  \item a screen will appear showing the video and a slider to indicate the valence and arousal value ranging in [-1,1]
  \item User annotate the video by moving the joystick up and down.
  \item File will be generated including annotation values and the corresponding time instances when the annotation value is generated.
\end{itemize}

\subsubsection{Preprocessing}
After we finishing annotating for videos, the next thing I need to do is preprocessing the videos. Firstly, there shouldn't be any useless frame in the dataset, useless frame means the frame that doesn't contain any facial images (like advertising and banners). After downloaded the videos, I have edited them manually using Format Factory, if there is a useless frame in the videos, I simply split the videos into two parts by cutting out the frame and make sure both parts are continuous clips.
Then, the dataset should be images instead of videos. I decided to use Menpo
software \cite{menpo} to extract the frames from the videos. Finally, I should use a detector to detect the face in each frame as we  want to focus on the facial expression and ignore the noisy part. There are several different tools I can use, I decide to go with the method described in \cite{detector}.  

\subsection{Existed Database}
Facial expression has been a popular subject in the last decade, there are many publicly available databases which can provide different images as dataset for training and research purpose.  Early database like Japanese Female Facial Expressions (JAFFE)\cite{JAFFE} and Multimedia Understanding Group (MUG)\cite{MUG} were captured under controlled environment, which means facial expression of people are posed. Researchers believe that the posed expression will be different from the natural expression with the same emotion, so the community started to focus on capturing the spontaneous expression. The most popular way to capture it is recording the reaction of people watching videos intended to elicit spontaneous emotion expression.Such as DISFA which recorded Twenty-seven young adults' reaction while watching such kind of video clips. \cite{DISFA} However, the limitations of this kind of database are the small amount of subjects (people),the same environment for every subject (recording environment) and also the same filming angle.

To address this problem and construct an in-the-wild database with different environment, A.Dhall built an database named Acted Facial Expressions in the Wild  (AFEW).\cite{AFEW}  It is a dynamic temporal facial expressions data corpus consisting of close to real world environment extracted from movies.\cite{AFEW} The database covers unconstrained facial expressions, varied head poses and movements, vast age range between 1 to 71, occlusions, varied focus, multiple people in the same scene and close-to-real world illumination. The limitation of this database is it only contains 700 images with 95 subjects in total.

Recola database is also a famous database that used in-the-wild data. It contains 9.5 hours of audio and videos about a 46 French people solving task in group.They expressed their natural behaviour and facial expression during this process.\cite{Recola}

SEWA database recorded volunteers' facial expression through a webcam while they watch adverts they provided by SEWA team. The volunteers fill in a questionnaire  about their emotion state during the watching process. This is also considered as a database that collected the data in-the-wild\cite{SEWA}.

However,most of annotated databases of facial expression in the wild are using category emotion model. Few of them used dimensional emotion model. To address this issue, AffectNet was created.\cite{AffectNet} AffectNet contains more than 1 million facial images collected from the Internet, It is constructed by searching images on three search engines using emotion related keywords. To annotate the Valence-arousal for the images, they hired 12 annotators to complete this task. This database is the largest  categorical and dimensional models of facial expression in the wild.\cite{AffectNet}

Aff-wild \footnote{recently Aff-Wild2 \cite{Aff-wild2,Aff-wild21,Aff-wild22} was released that contains annotations for valence-arousal, action units and basic expressions} is another database that used dimensional model to annotate the in-the-wild facial images . It was created by Dimitrios Kollias and his teammates. The data was collected from video-sharing website like Youtube and there are 298 videos displaying the reaction of 200 subjects.The search keyword is "reaction" and the database displays subjects reacting to a variety of stimuli. The database was annotated using a joystick and Valence and Arousal range continuously in [-1,+1].\cite{aff-wild1}

This is a table to compare some of  the famous existed database.
\begin{center}
 \tiny
 \begin{tabular}{||c c c c||} 
 \hline
 Database & Number of images/videos & Resolution & Type \\ [0.5ex] 
 \hline\hline
 JAFFE & 213  & 256* 256 & Posed \\ 
 \hline
 MMI Database & 1280 videos4 & 720* 576 & Posed and Spontaneous \\
 \hline
  (RaFD) & 8040 images & 	681*1024 & Posed\\
 \hline
 Oulu CASIA NIR VIS database  & 2880 video  & 	320x240 & Posed \\
 \hline
 Belfast Database & 180 video clips & 1920*1080 & Natural Emotion \\
 \hline
 ISED & 428 videos  & 1920* 1080 & Spontaneous \\
 \hline
 AffectNet & 950000 & Various & Wild setting \\
 \hline
  Aff-Wild & 1250000 & 640x360 & In-the-Wild setting4 \\
 \hline
 
\end{tabular}
\end{center}

Because the purpose of this project is to build the model using 4k resolution images, and there is no existed publicly database that contains data with 4k resolution, I didn't choose to use any of those existed database. However, I build my own database using the video from Youtube. All the facial expression are captured in the daily life with most of them 4k resolution. They will be annotated using valence-arousal model. The details of the database will be mentioned in the later of the article.

\subsection{Artificial neural network}
The computational Neural network model was created by Warren McCulloch and Walter Pitts based on algorithms called threshold logic. It's consisted of a large amount of artificial neurons which take an input and generate output depends on the input and activation. 
Neural network is considered as a function or a distribution, and this function is a composition of other functions. Each artificial neuron represents a function and the dependency between those neurons formed the structure of the neural network\cite{create of neural network}.

\begin{figure}[h]
	\centering
    \includegraphics[width={0.6\textwidth},height={0.3\textwidth}]{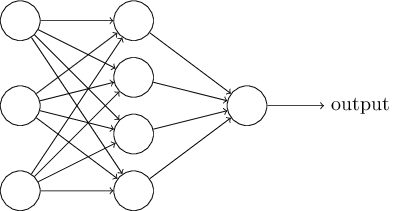}
    \caption{Neural Network}
    \label{fig:clean_curve}
\end{figure}

The connection (dependency) between neurons are represented as a transferring from a output of neuron $i$ (successor) to a input neuron $j$ (predecessor) :$P_j(t) = \sum_{i} o_i(t)$. The input of neuron $j$ (successor) is equal to the sum of all the output value of neurons $i$ (predecessor). However,  the connections are assigned to weights and bias as well. Which means $P_j(t) = 
\sum_{i} o_i(t)w_{ij}+w_{0j}$ , $w_{ij}$ is the weight on the 
connection from $i$ to $j$, $w_{0j}$ is a basis and this is called propagation function.\cite{propagation} The learning process of neural network can be considered as modifying the weights and threshold until it find the favoured solution to the problem.

\subsubsection{activation function}
However, the original neural network can only represent linear function and the complexity of the model will be limited. To solve the complex task like image recognition, the non-linear function is required. So scientist used another function to redefine the output of each neurons , and this function is called activation function. The neural network can present non-linear function if the activation function is non-linear.\cite{what is activation function}

 There are large amount of activation functions can be chosen at different layers. And the choice depends on the properties of the activation function. Those are the property we may need to consider while choosing the function:
 
\begin{itemize}
  \item Nonlinear: A two-layer neural network model can be proven to be a universal function approximator if the activation function is non-linear. \cite{nonLinear}
  
  \item Range: it's preferable to use activation function with finite range if the pattern presentations are designed to affect only limited weights and infinite range functions are favoured if it aim to affect majorities of weights.
  
  \item Continuously differentiable: This property is required for gradient-based method. Non continuously differentiable function could leads problem in such method.\cite{Differentiable}
  
  \item Monotonic:When the activation function is monotonic, the error surface related with a single-layer model is convex. \cite{monotonic}

\end{itemize}

\subsubsection{learning process and optimisation}
The learning process of neural network can be separated into two phases. The first part is to propagate forward through the network and work out the result $y_i$, this process is called feed forward. Then calculate the difference between $y_i$ and  the expected output $y$. Feed the difference (error) back through the neural network. Calculate the delta(error) for every nodes on the network. it's called back propagation.

The second part is weights updating. For every weights, the weights output and delta will find the gradient of the weights. And the new weights will be generated by subtract a ratio $r$ of the gradient. This is called gradient descent and $r$ is named learning rate. The purpose of the gradient descent is to find the global minimum of the loss function which will leads the minimum error. The larger learning rate can speed up the converging process. But it can also cause problem of overshoot. To improve the learning in terms of both performance and speed, an adaptive learning rate can be useful. \cite{learning_rate}.
\begin{figure}[h]
	\centering
    \includegraphics[width={0.6\textwidth},height={0.3\textwidth}]{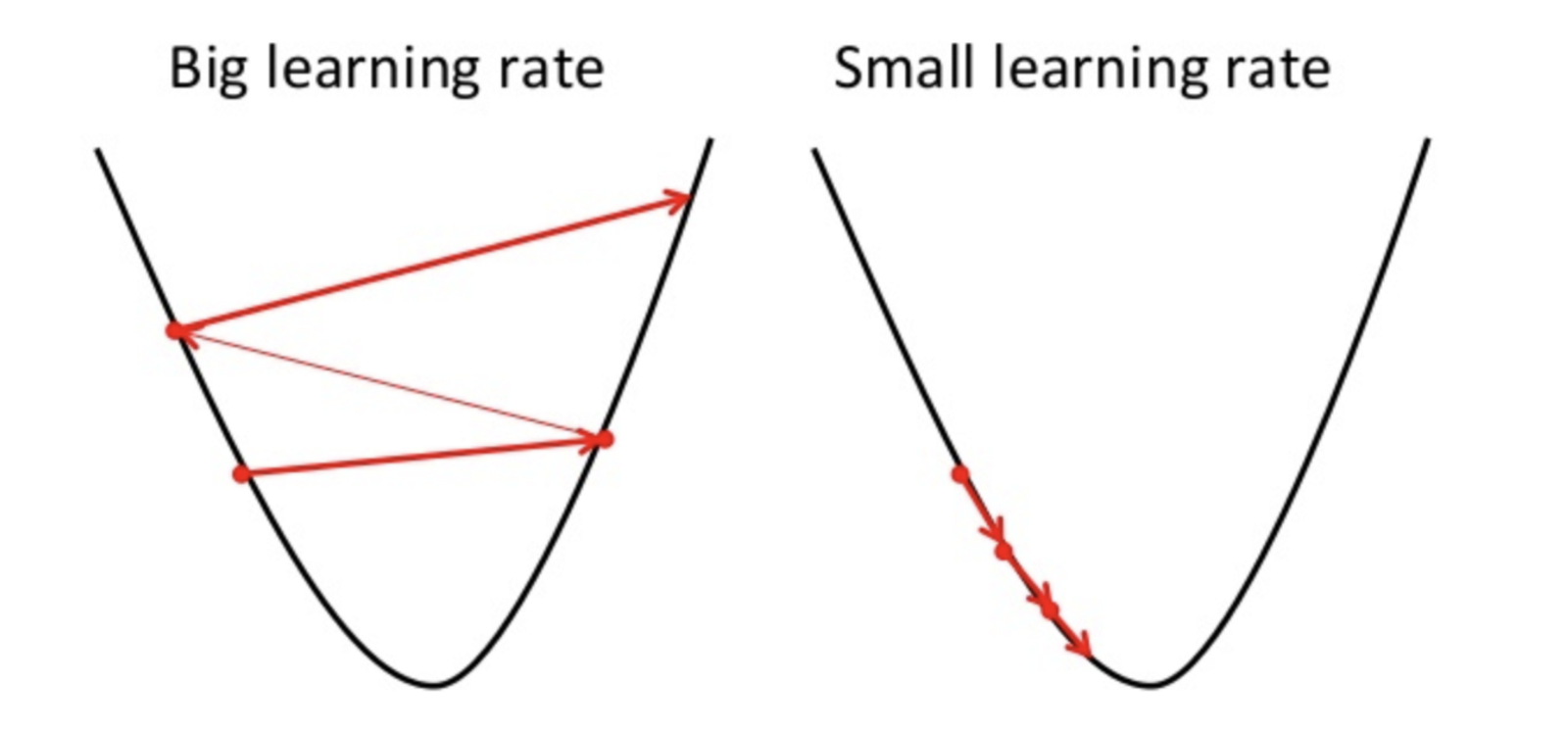}
    \caption{Neural Network}
    \label{fig:clean_curve}
\end{figure}

The learning processes will be repeated until a certain point. Normally the point will be set manually as a certain amount of epochs (complete presentation of the data set has been fed into network) finished. But it's possible for the people to perform an early stop.

This is a brief introduction of neural network. In the next subsection, I will talk about a famous neural network model called Convolutional neural network, and it's the most commonly used model for emotion recognition.

\subsection{Convolutional Neural Network}
Convolutional neural network is a type of deep neural network model and it's wildly applied to analyse images. It's inspired by the connectivity pattern between neurons resembles the organisation of the animal visual cortex \cite{cnn biology}.

A Convolutional neural network (CNN) contains one input layer and one output layer along with a lot hidden layers. The hidden layers are consisted of convolutional layers, RELU layers, pooling layers, fully connected layers and normalisation layers. \cite{CNN}

Neural network (fully connected neural network) can be used to learn feature and classify the input, however if we want to classify a image, the size and architecture of the input will become a problem. The large size of image will lead to a large number of neurons and weights. For a image with 100*100 resolution, we will need a network with 10000 weights for each neurons in the second layer. The complexity of the model could cause problem like gradient vanishing or exploding.  Convolutional layer resolved this problem by applying a filter to reduce the number of free parameters in of an image.\cite{Convolutional layer} The image below is an example, the input image has size of 8*8 pixels, the filter is 3*3, filter will slide through the input data, and output an image with 6*6 pixels. We can repeat this process for larger size image. 

\begin{figure}[h]
	\centering
    \includegraphics[width={0.6\textwidth},height={0.3\textwidth}]{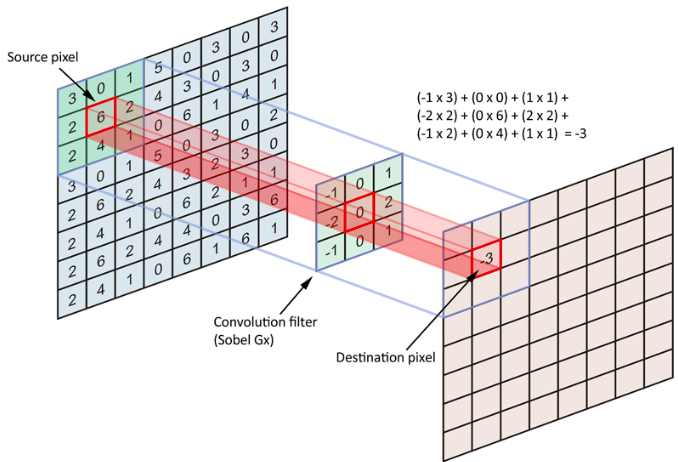}
    \caption{Convolutional layer}
    \label{fig:clean_curve}
\end{figure}

Pooling layer can combine the neurons output in one layer into a single neurons in the next layer, one common pooling method is called max pooling which select the maximum neuron with the maximum value.  Fully connected layer is same as feed-forward layer which is used to learn to classify the data. 

The full learning process of convolutional neural network is as follow.

\begin{figure}[h]
	\centering
    \includegraphics[width={0.8\textwidth},height={0.3\textwidth}]{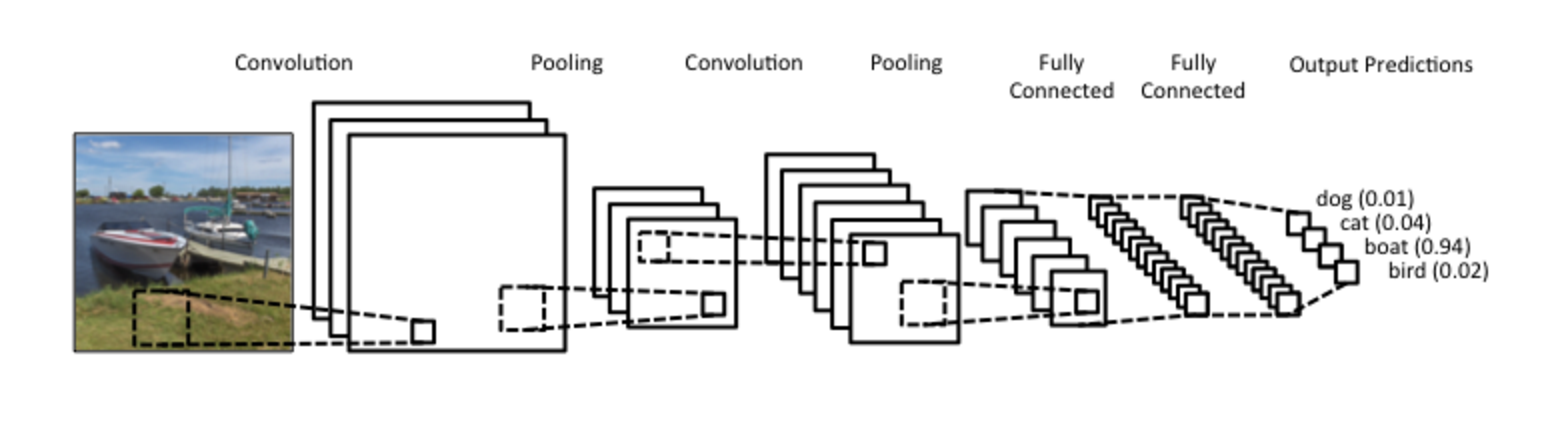}
    \caption{Convolutional neural network}
    \label{fig:clean_curve}
\end{figure}

Convolutional neural network has been wildly used for image recognition, the main advantage of CNN is that it can highly removed the dependence on physics-based models and other preprocessing techniques by enabling “end-to-end” learning directly from input images\cite{advantage_CNN}. Most methods for emotion recognition are using CNN directly, some of them are using a modified version of it.

Breuer and Kimmel  applied CNN visualisation techniques to understand a model learned using various facial emotion recognition (FER) datasets \cite{Breuer}. It represented the capability of networks trained on emotion detection, across both datasets and various FER-related tasks. Jung et al combined two different CNN-based methods together to improve the performance. \cite{Jung} First model extracted temporal appearance features from the image sequences. Second model extracts temporal geometry features from temporal facial landmark points.

However, there is one limitation of CNN which is that it can't reflect the temporal variations in the facial components, in another words, it can't be useful of analysing a sequence of frames cut out from a  video. To meet this requirement,  researchers introduced a new hybrid type which combine CNN with long short-term memory (LSTM) which is a special type of Recurrent Neural Network (RNN) capable of learning  long term dependencies. This structure enables the network to generate output based on the desired earlier outputs.\cite{cnn_rnn}

\begin{figure}[h]
	\centering
    \includegraphics[width={0.8\textwidth},height={0.5\textwidth}]{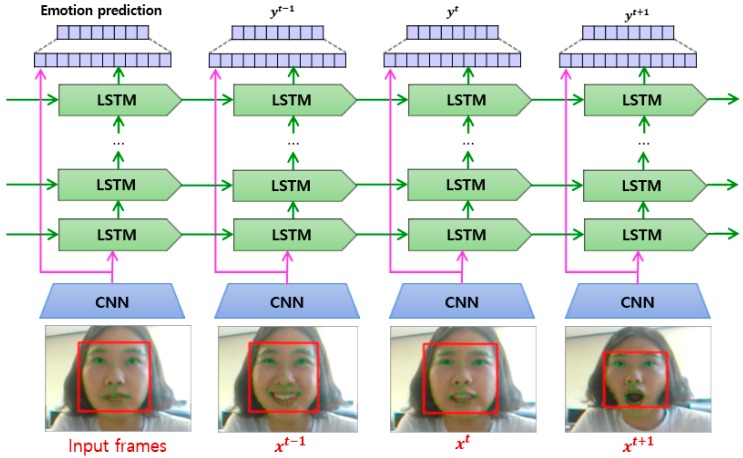}
    \caption{CNN with LSTM}
    \label{fig:clean_curve }
\end{figure}

Kahou et al. \cite{Kahou} proposed a hybrid RNN-CNN framework for propagating information over a serials using a continuously valued hidden-layer representation. They presented a complete system for the 2015 Emotion Recognition in the Wild (EmotiW) Challenge , and proved that a hybrid CNN-RNN architecture for a facial expression analysis can outperform a previously applied CNN approach using temporal averaging for aggregation.

\newpage

\section{Build the database}
The main purpose of this project is to train the neural network using 4k resolution images that are captured under random situations which is refered as  in-the-wild. However, there is no such kind of public database that satisfied my purpose, so I have constructed my own database.

\subsection{Download Youtube videos}
There are two ways to construct the database, either collecting a millions of images or collecting hundreds of videos and cut them into frames. I believe emotion is something varying continuously and a sequence of frames will be better for neural network to learn compare to independent static images.So I decided to collect videos instead of images. Youtube is the best website for me to download the videos, not only because of the large quantity but also the searching functionality it has is going to be helpful during my searching process.

The category of videos are important for this process, have a right category can save a lot of time by using the searching engine efficiently. To decide what type of videos I am looking for, here are some important factors I would need to consider about:
\begin{itemize}
\item Videos must have 4k quality. 
\item Videos must be filmed  under wild environment (not lab controlled environment). 
\item Videos should be filmed about the people, and the face of people should be clear to identify.
\item Videos should be continuous which means the angle of the camera and the content in the videos are continuous for at least 2 minutes (I only cut 2-3 minutes for each video). 
\end{itemize}

By considering the above factors, I found three types of video satisfiable: Vlog, Interview, and Reaction videos. So most of the videos I downloaded are either one of them. However, there is still one more thing I needed to consider, which is the diversity of emotion in my database. I need to keep the balance between the amount of videos with different emotions. So my strategy is to search the correct type of videos with certain emotional keywords which are from the 7 basic emotions ( anger, fear, disgust, happiness, sadness, surprise, contempt ) and manually control the number of the videos I downloaded for each emotion.

\begin{figure}[h]
	\centering
    \includegraphics[width={1.0\textwidth},height={0.4\textwidth}]{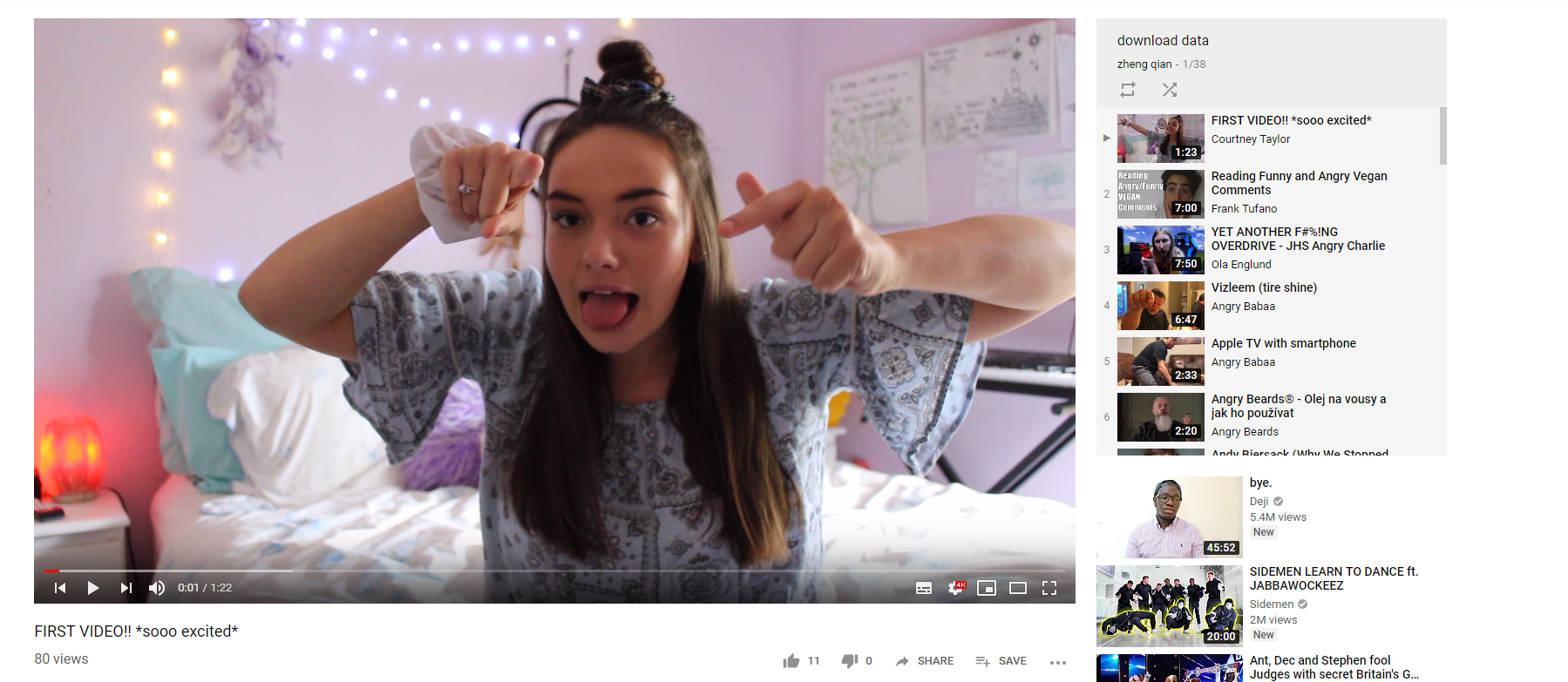}
    \caption{Youtube playlist}
    \label{fig:clean_curve }
\end{figure}

I added all the appropriate videos into my personal Youtube playlist and used Youtube-dl\cite{ youtube-dl} to download all of them with 4k resolution quality. Because some videos are really long, I  need to cut them into 2-3 minutes manually.Also all the videos have  different frame rates, choosing a fixed frame rate for them will be helpful during the annotation. I wrote a little program to check the frame rate for all the videos, and according to the result, I noticed most of the videos have 25 FPS. So I  used Format Factory\cite{format-factory} to convert them into an AVI format with 25 frame rate and cut out a 2-3 minutes part for each one of them.

\begin{figure}[h]
	\centering
    \includegraphics[width={0.4\textwidth},height={0.2\textwidth}]{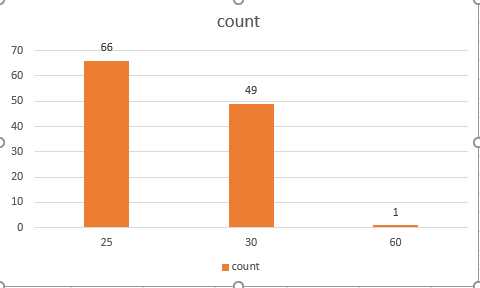}
    \caption{Youtube playlist}
    \label{fig:clean_curve }
\end{figure}

There are 116 videos which contain 403040 frames in total and it should be enough  for training and testing, so the next step is to annotate the videos.

\subsection{Annotation}
Annotation means to  label the data set, in my case, the data set is a bunch of frames and the labels are arousal and valence values since I am using dimensional emotion representation for emotion prediction.The values are going to be between -1 and 1. For arousal values, 1 means extremely active and -1 means extremely passive, and for valence values, 1 means extremely positive and -1 means extremely negative. I have to give each frame a arousal value and a valence value from my own understanding of the emotion the people has at that frame. However, there are two issues I need to solve for annotation: 

1. There are 403040 frames in total, it will take a lot of time to simply record the values for each frame, I needed to use a tool to save time. 

2. The values annotator is going to label completely depends on annotator's judgement, however human have bias on reading emotion. So instead of doing the annotations by myself, I need to find a group of annotators to do it with me and I can get rid of the bias by calculating out the average. 

\subsubsection{annotation application}
To solve the first issue, luckily I have get right to use the annotation application that built by Dimitrios and his team. This application is designed to continuously annotate the videos with valence and arousal values. It doesn't allow the annotator to judge both valence and arousal values at the same time because it will be hard to reach high quality on both of them. Instead, annotator has to select which values to annotate  first. Then annotator will need a joystick to do the annotation, if the application detected the joystick, then the annotator can start to annotate the videos that has been saved in the demanded folder\cite{Aff-wild}.

Once the annotator started the application, there will be a video window  and a bar on the right hand side of it that indicate the valence or arousal value you decided to label at this frame. Annotator use the joystick to change the value by pushing up or pulling down and he can choose to play or pause the video. Once the video start to play, the value that indicated on the bar will be signed continuously to each frame. When  one video finished playing, the application will generate a file with each line contains the time frame and the value assigned to that frame.

\begin{figure}[h]
	\centering
    \includegraphics[width={0.6\textwidth},height={0.4\textwidth}]{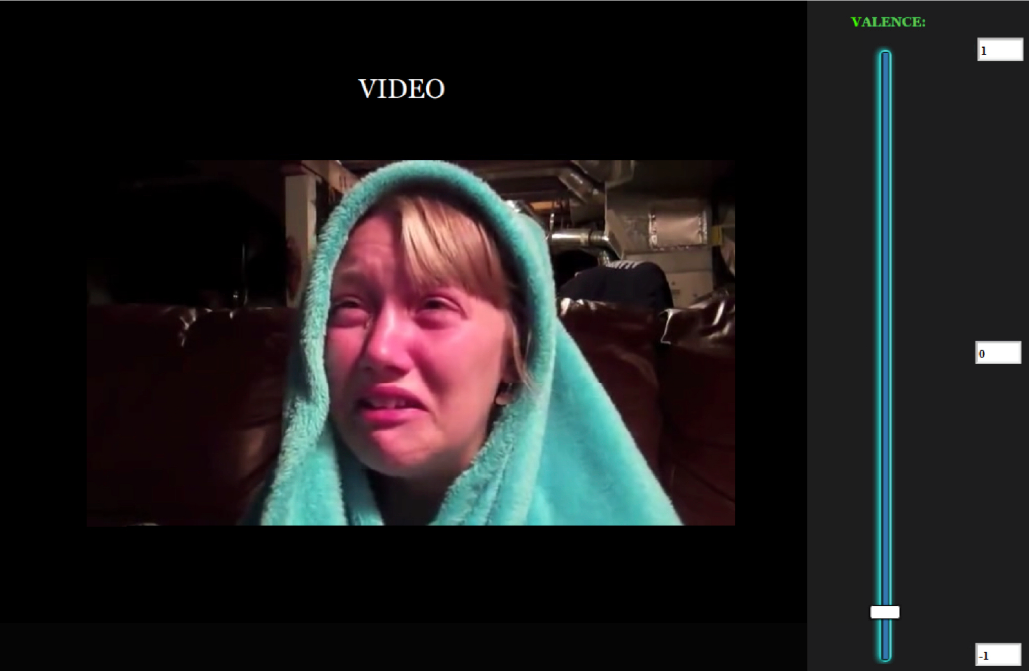}
    \caption{Annotation application UI}
    \label{fig:clean_curve }
\end{figure}
 \newpage
 
 \subsubsection{Annotation processing}
 To solve the second issue, Dimitrios' team members helped me with it, there are 4 professional annotators in the team and they annotated all the videos using the application I mentioned above. After collecting the results, we have to make sure the quality of those result by calculating out the inter annotation correlation: 
 \newline
 
\begin{table}[h]
    \centering
\begin{tabular}{lllll}
      & ann 1 & ann 2 & ann 3 & ann 4 \\
ann 1 &       & 0.875 & 0.603 & 0.875 \\
ann 2 & 0.875 &       & 0.667 & 0.784 \\
ann 3 & 0.603 & 0.667 &       & 0.651 \\
ann 4 & 0.875 & 0.784 & 0.651 &      
\end{tabular}
\end{table}

From the correlation table, we can tell the result is high, the average correlation is 0.74 which means the annotation processes are consistent among those professional annotators, and the results are reliable. So for each frame, we calculated the average arousal values and valence values among the 4 annotators' results. Finally, I split the videos into frames, and save them into different folders. Generated a file with the following format : Frame photo path, valence value, arousal value. At this point, the database has been constructed.

\begin{figure}[h]
	\centering
    \includegraphics[width={0.6\textwidth},height={0.4\textwidth}]{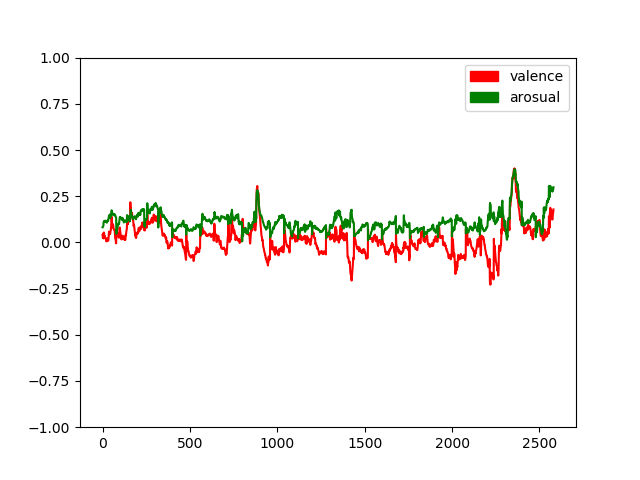}
    \caption{Valence and Arousal result example}
    \label{fig:clean_curve }
\end{figure}
 \newpage
 
\section{Facial detection and alignment}
Before we input the images into the network, we need to preprocess those data. One of the most important task to do is face detection. It is designed to speed up the training process and increase the recognition accuracy by detecting the features of face from a image and cropping the image into a fixed mask size. The idea behind it is to keep neural network focusing on the face features instead of irrelative object like body or feet.

In this project, I am using Multi-task Cascaded Convolutional Networks to perform the face detection. It uses three different CNN to do three different tasks. Firstly, it produce a candidate window for image, then it refine the window by removing the non-face object (like arm, hand). Finally it refine the output again to produce 5 facial landmarks.\cite{face_detection} In my case, I resized each image into a $112 \times 112 $ bounded window.

Another important process included in the face detection is called alignment. It produced a line at the centre of face from top to bottom for each image,after the face detection and cropping, it will rotate the output image to ensure the line is perpendicular to the bottom bound.

The purpose of this process is to make sure no matter what original position and angle of the head in the image is, all the output will end up in a same position and angle. This is really helpful to my database because the data I have are in-the-wild, which means people have different poses and firming angle. Those difference could lead to a big error without alignment.

\begin{figure}[h]
	\centering
    \includegraphics[width={0.8\textwidth},height={0.4\textwidth}]{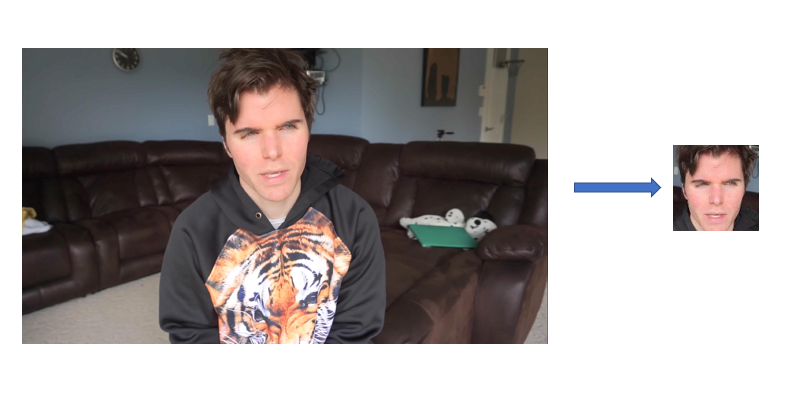}
    \caption{Face detection}
    \label{fig:clean_curve }
\end{figure}

\newpage
\section{Neural network}
Neural network has been applied to classify the emotion from facial expression for a long time, there are already lots of famous models that could be suitable for my task.  However, to choose the one that suit our task the most, I have to train all of those good models with my data set and find out the best one according to the evaluated result. In this section, I will introduce the models I have tested on and worked well, all of them are the models of Convolutional neural network since CNN is the best network to process images (I have talked about it in the background).However, instead of using simple CNN structure, I modified them into CNN-RNN architecture, and I will get into details about it as the last model. Furthermore, I am using a loss function called Concordance Correlation Coefficient (CCC). I will talk about it after the models' section. And in the final subsection, I will give the evaluations (results) for all the models I have talked about so that I can make a comparison to decide which one is the best for my project.

\subsection{Alexnet}
Alexnet is a type of convolutional neural network designed by Alex  
Krizhevsky \cite{Alexnet},the original network was expensive to compute because of the large depth of network, however the utilisation of GPU conquered this problem.Alexnet is designed to train with large size of data set and high resolution images,and it is considered one of the most influential papers published in computer vision, having spurred many more papers published employing CNNs and GPUs to accelerate deep learning\cite{Alexnet_influnce}.

From the graph about the architecture of the Alexnet, we can notice that it is formed with a series of convolutional layers, and  each layer is followed by a Max pooling layer, a Relu activation layer and a normalisation layer. The first convolutional layer filters the $224 \times 224 \times 3$ input image with $11\times11\times3$ 96 kernels, the second layer use 256 $5\times5\times 48$ kernels . The third layer has 384 kernels have the shape of $3\times3\times256$.The forth layer uses 384 $3 \times 3\times192$ kernels and the fifth layer use 256 kernels has the shape of $3\times3\times192$. 

\begin{figure}[h]
	\centering
    \includegraphics[width={0.8\textwidth},height={0.4\textwidth}]{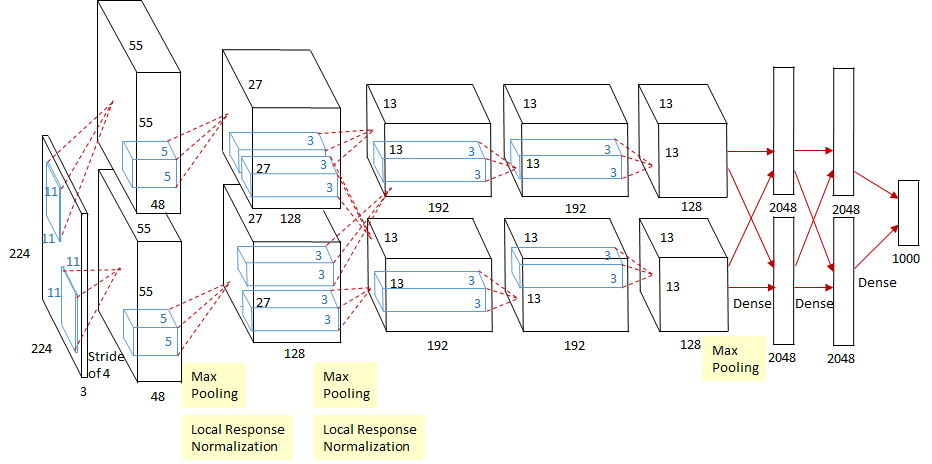}
    \caption{Alexnet}
    \label{fig:clean_curve }
\end{figure}
\newpage
\begin{figure}[h]
	\centering
    \includegraphics[width={0.5\textwidth},height={0.5\textwidth}]{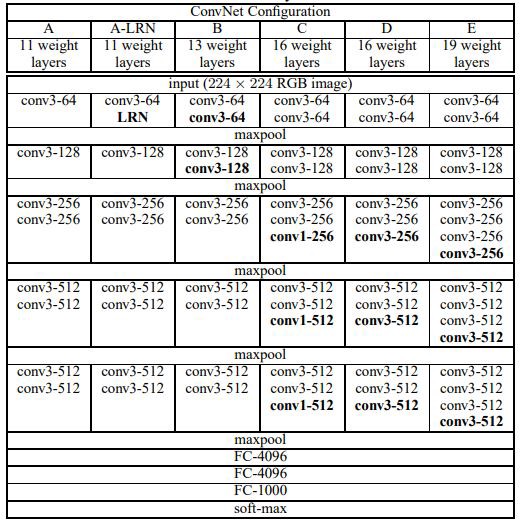}
    \caption{Alexnet layers}
    \label{fig:clean_curve }
\end{figure}
In the original model, the output of convolutional layers  have the size of $12 \times 12 \times 256$, however, the size of input images I used from my database is $96 \times 96 \times 3$, the size of output images become $4 \times 4 \times 256$. 

\subsection{VGG16}

VGG16 is a Convolutional neural network model proposed by K. Simonyan and A. Zisserman. \cite{VGG} The model achieves 92.7\% top-5 test accuracy in ImageNet, which is a dataset of over 14 million images belonging to 1000 classes \cite{VGG_achievement}. It improved the performance of Alexnet by replacing the large kernel-sized filter with a series of $3 \times 3$ kernel-sized filters. As same as Alexnet, the network is decent at classifying the high resolution images, however the drawback is the training process take a long time. Luckily the department of computing in Imperial college provided a bunch of NVIDIA  Titan GPUs for me to use, with the power of calculation they have, the training duration is decreased to 20 hours only.

\begin{figure}[h]
	\centering
    \includegraphics[width={0.8\textwidth},height={0.4\textwidth}]{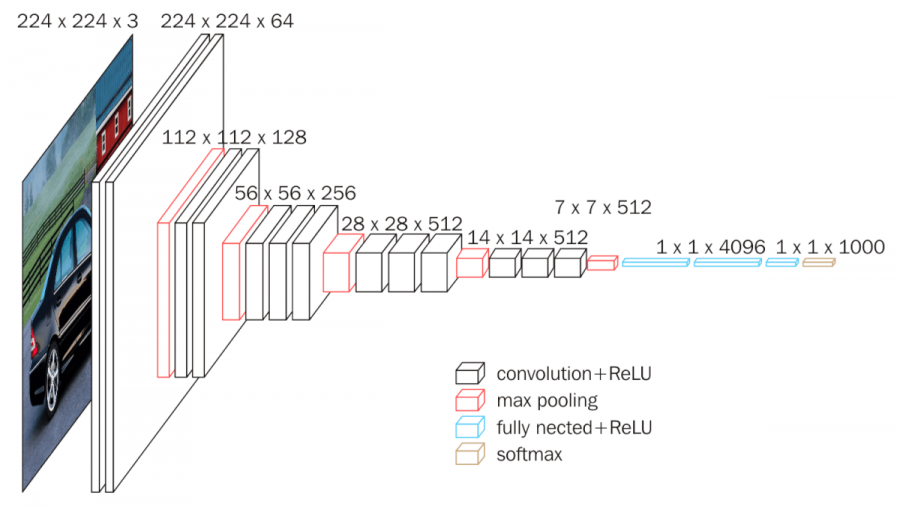}
    \caption{Vgg16}
    \label{fig:clean_curve }
\end{figure}

The input of the original VGG network is $224 \times 224$ RBG image. The image is past through a series of convolutional layers and max pooling layers. It uses $1 \times 1$ convolutional filter for every convolutional layers, fixed at 1 stride and the spatial padding of convolution. The resolution of image (input) will not be changed after the convolutional layer. However after every 3 convolutional layers, it performs a max - pooling on $2 \times 2$ filter window with stride 2 \cite{VGG_architecture}. 

There are three fully connected layers follow the stack of convolutional layers, the first two layers have  4096 channels each, and the last one have 1000 channels perform a 1000-way ILSVRC classification. Relu activation function is applied at each hidden layers.
\newpage

\begin{figure}[h]
	\centering
    \includegraphics[width={1.0\textwidth},height={1.0\textwidth}]{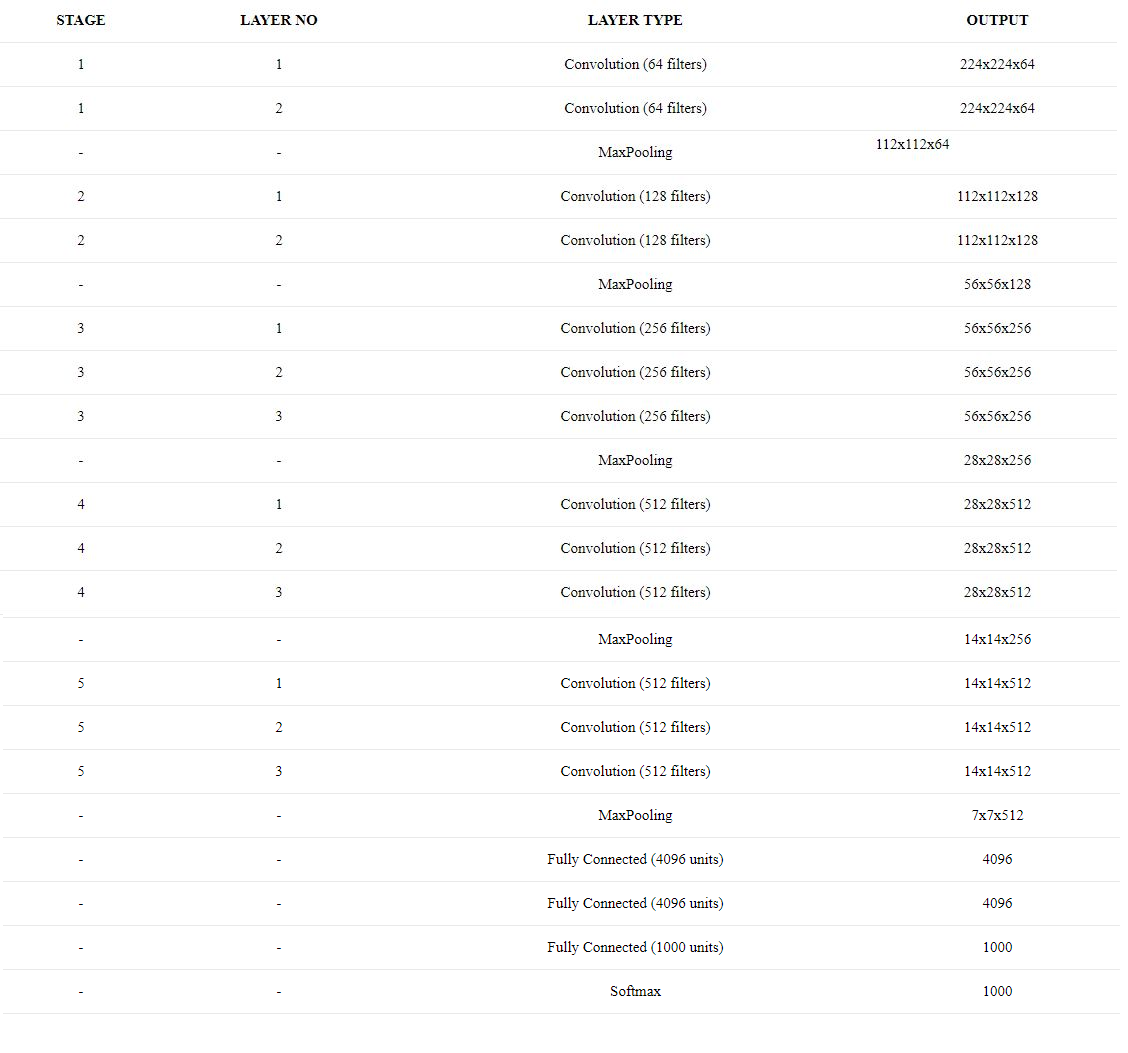}
    \caption{VGG16 layers}
    \label{fig:clean_curve }
\end{figure}
In original VGG16 network, the shape of output from stack of convolutional layers is $7 \times 7 \times 512$, however, because I was using $96 \times 96 \times 3$ images as input, the output size become $4 \times 4 \times 512$, and the output of full connected layers become 2 instead of 1000 because we are predicting arousal and valence values rather than 1000-way classification.

\subsection{ResNet}
A residual neural network (ResNet) is a type of artificial intelligence neural network. Residual neural networks utilize itself by skip connections, or short-cuts to jump over some layers\cite{RNN_WIKI}. One of the motivations for skipping the layers is to avoid the vanishing gradients,by reusing the activation function from the previous layers. Another reason is to simplify the neural network architecture. 

\subsubsection{degradation problem}
We believe that when we increase a neural network's depth, the accuracy gets saturated because we believe a deep network should model the intricacies data well and when a more powerful network is provided with additional layers, it should completely model out our data. However in reality, it's not the case. The training error increased when we added additional layers onto a neural network. And this is called degradation problem.\cite{Resnet_paper}

However, intuitively the error of deeper neural network should be at least the same as the shallow one. Let's say if we have m layers neural network A, and another network B  with n layers where $m > n$. If for first n layers in A, it has exactly weight and structure as the network B, and the rest of the layers in A are all identity layers (the output is the same as input), then the output of A should be the same as B.Thus, A can easily learn B's representation, and  if the data are more complicated than this, A is expected to learn it.
\subsubsection{residual block}

\begin{figure}[h]
	\centering
    \includegraphics[width={0.6\textwidth},height={0.3\textwidth}]{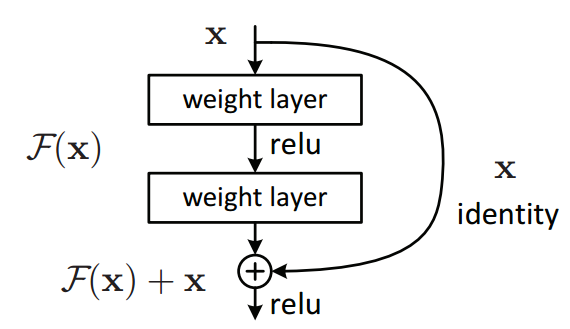}
    \caption{Residual block}
    \label{fig:clean_curve }
\end{figure}
The solution of the degradation problem is to learn identity function of the layers, from  the  graph, we can see the output of weight layer is fed directly into the next layer and also the layer one hop away, so it can directly learn the identity function by relying on skip connections only. And the residual block solved the degradation problem for us.

\subsubsection{architecture}

The ResNet is composed of a series of residual block, because of the solution to degradation problem I mentioned above, ResNet has much more hidden layers compare to other CNN structure like VGG16. It has one convolution layer with $7 \times 7$ filter, and all the rests are using $3 \times 3$ filters.  This is a graph for the detailed structure.
\newpage
\begin{figure}[h]
	\centering
    \includegraphics[width={0.6\textwidth},height={1.0\textwidth}]{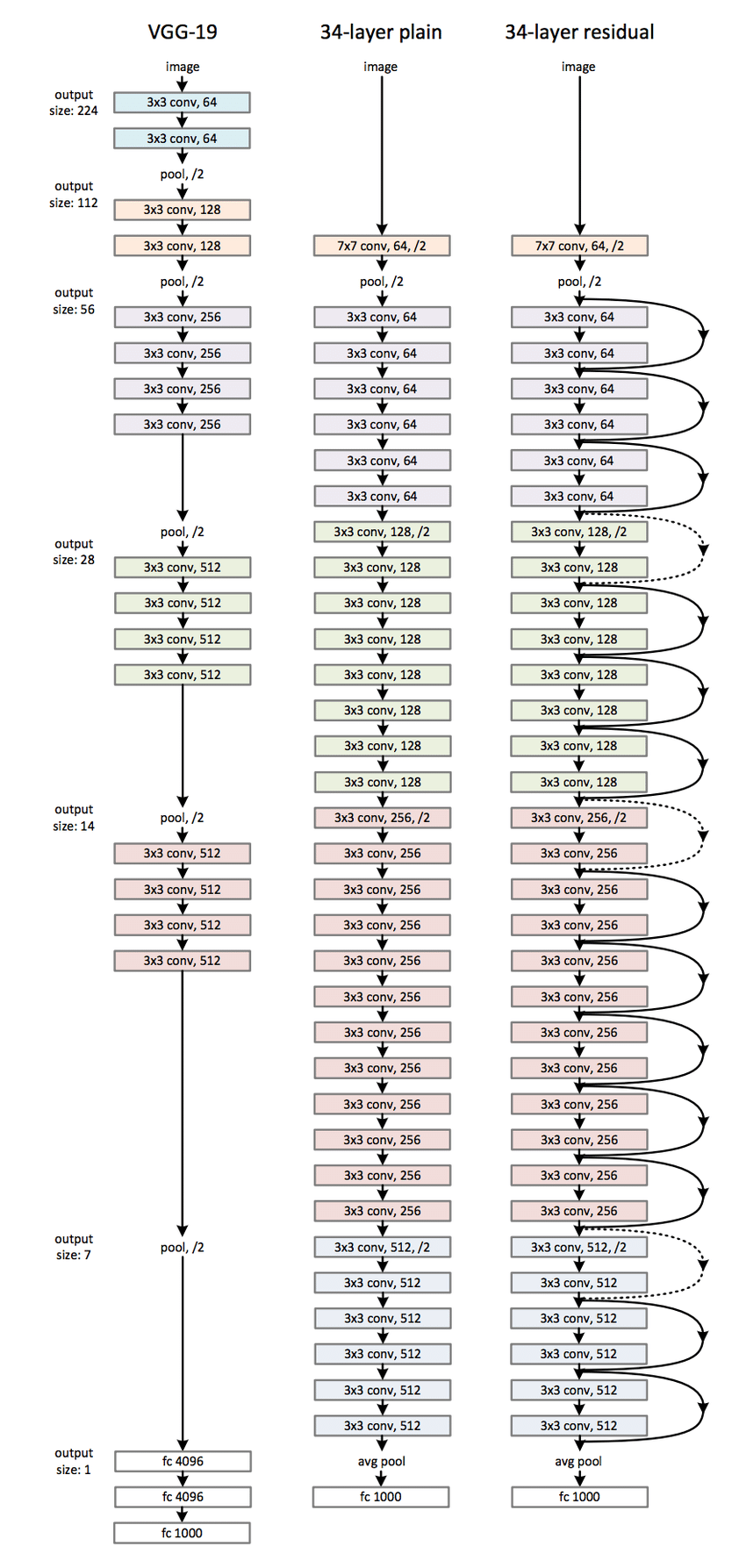}
    \caption{ResNet}
    \label{fig:clean_curve }
\end{figure}

\subsection{CNN-RNN structure}
The database contains the images that cut from a series of videos, which means the images from the same video are continuous. I believe people's emotion change smoothly and gradually,and the arousal and valence values should be similar between two continuous frame.Thus, learning from the image itself, and also the images it's following and followed, should improve the performance of the network. 

Instead of only train with CNN model, I decided to use the hybrid CNN RNN deep neural architectures as well. The idea of the CNN-RNN structure is to flatten the latent space in CNN network, then feed the result into the RNN units, and the RNN units will predict out the result.

\noindent An existed example is VGG16-GRU network created by Kollias Dimitrios.\cite{vggface_gru_structure,multi-component}

\begin{figure}[h]
	\centering
    \includegraphics[width={0.4\textwidth},height={0.6\textwidth}]{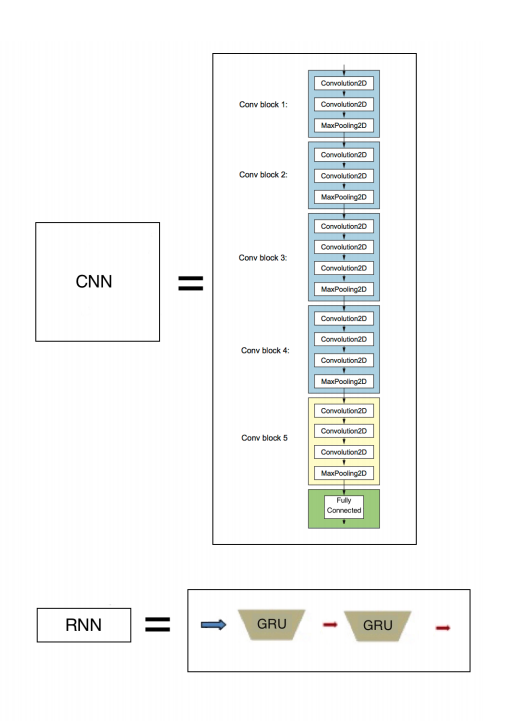}
    \caption{VGG16\_GRU}
    \label{fig:clean_curve }
\end{figure}

Compare to VGG16, The CNN part of VGG16-GRU is the same.And It is followed by a fully connected layer which flatten the latent space into a $4096 \times 4096$ vector. On top of that, a 2 layered RNN is stacked, which take the vector as input and give the final estimates for valence and arousal.\cite{vggface_gru_structure}

Follow the same pattern, I have implemented the Alexnet-GRU and ResNet-GRU by get ridding of the last full connected layer of them, feed the output into the RNN units. In this project, I have tested both CNN model and the CNN-RNN type model, the result shows CNN-RNN architecture outperform the CNN model with my database.

\subsection{Loss function and evaluation}
The network is supposed to predict two values for each image, and calculate out the loss by comparing with the labelled values on the image. Usually, MSE (mean squared error) would be used as loss function in this case, however, I  used  1 - CCC (Concordance Correlation Coefficient) as the loss function instead. In static, CCC evaluate reproducibility or for inter-rater reliability. Which means it is a score of how much homogeneity or consensus exists in the ratings given by various judges. \cite{CCC} CCC is defined as follow:

{\centering
  $p_c = \frac{2{S_y}_{f(x)}}{{s^2}_{f(x)} + s^2_y + (\overline{f(x)}-\overline{y})^2}$\par
}

${s}_{f(x)}$ and $\overline{f(x)}$ are the variance and mean of the predicted values. $s_y$ and $\overline{y}$ are the variance and mean of the corresponding labelled values.with $x$ being the input sequence of the network. ${S_y}_{f(x)}$ is the covariance.

I compute the mean and variance first for each sequence,  then I use those values to compute CCC. The CCC calculated the MSE in $(\overline{f(x)}-\overline{y})^2$ part, when the MSE decrease, the CCC will increase. However it also evaluated the correlation between the labelled values and predicted values. 

The reason I take correlation into account is because I are dealing with frames from videos, the emotion change through the videos should be correlated between the labelled values and predicted values. Thus, I want  the MSE to be small and the correlation to be big . Which lead us to the conclusion: a good network should give the predictions with high CCC and this is how I decided to evaluate the results. However, because I prefer to perform gradient descent than gradient ascent, I set the loss function to 1 - CCC instead of CCC.
\newpage
\section{Experiment}
In this project, I planned to use the database that I built in order to train different models of the neural networks that I mentioned above. By working with different hyperparameters I could find the optimal result for each model. By comparing those optimal results, I should be able to come to a conclusion about the performance of the models on the dataset that I have created for this project. The training scripts, testing scripts and all the neural networks model were all implemented using the Tensorflow framework. \cite{tensorflow}. The reason I chose Tensorflow instead of other framework is becasue, 1. I am really familiar with it and have already worked with it on several other projects. 2. Tensorflow has so many existed model for me to use so that I can turn them into the CNN-RNN model I need for my project.

All the models were trained on two different GPUs. The first one is the Nvidia GeForce GTX Titan X and the second one being the Nvidia GeForce GTX 1080 Ti which were provided by the department of computing of Imperial College.

\subsection{Preprocessing}
There are three different preprocessing methods that can be used in this project:

\begin{enumerate}
    \item Normalisation: normalises the value of each pixel into a range between -1 to 1.
    \item Mean Subtraction: it subtracts the mean of all the pixel values from every pixel in the training set.
    \item Whitening: it first performs mean subtraction, then divides every pixel value by the standard deviation of all the pixel values in the training set.
\end{enumerate}

The type of images in my input data are RGB images, which means for each pixel (element) in the image, the value will be a number between -255 to 255. However, the valence value and arousal value the network is supposed to predict is between -1 and 1. Obviously it's hard to generate these results with a large input value, so normalisation is an important process before I feed the input into the neural network. The method is: for each value in the pixel, minus 128 and divided by 128 (128 is half of 255) and every values will be in the range between -1 to 1.

Mean subtraction and whitening require me to calculate out the mean $\mu$ and variance $\sigma$ of the whole training set. Mean subtraction involves subtracting every value by $\mu$. Whitening is when values are subtracted by $\mu$ and divided by the root of $\sigma$. These are two famous pre-processing methods. Mean subtraction is performed in the VGG16 object detection model that I have downloaded.

\subsection{Data reader}
All the images in my database are saved in a folder on the DOC system. As I mentioned before, I generated a file that saves the path for each image along with the valence and arousal value that corresponds to it.

I used the data reader from the Aff-Wild evaluation model to read this file. The reader will go through the file line by line and put them into a big list. The list is then divided into a number of sub-arrays. The elements of the sub-arrays will have the format  as {image path, valence value, arousal value}. Each sub-array  has the fixed sequence length of 80 which is chosen by me since 80 is the sequence length chosen by Dimitrios for Aff-wild model as well. The sequence length is the length of sequential data that is inputted into a Recurrent neural network. Each of the sub-arrays in the list is called a batch.

I then divide the list into a training set and test set which partitions into a ratio of roughly 2:1. The training sets end up with 268640 training samples (3358 batches) and 134400 test samples (1680 batches). Then I use slice\_input\_producer from Tensorflow framework to extract n of them out every time I input them into the network. In this project $n$ must be bigger than 1 and the list must be shuffled by the unit of the batch, so the n batches that extracted out are less likely to be in the same video. The reason for this is that for the 1-CCC loss function I am using, if the images are all from the same video, the covariance will be really small. Hence, the CCC is likely to be 0 or nan, which will be harder for the network to learn from. However, if there is more than one batch and the batches are from different videos, then the variance will be of large enough to avoid the CCC becoming b 0 or nan. $n$ is chosen to be 4 in the beginning, however I tried with different $n$ and 4 did give me the best result out of them.

\subsection{Hyperparameter}
I chose to use Adam Optimizer with an initial learning rate of $1^{-5}$. The learning rate is provided by the VGG model and I have tried with other learning rates. The model won't learn anything with a learning rate as $1 ^ {-4}$ and model will converge slowly if learning rate is $1 ^{-6}$. So $1 ^{-5}$ seems like a good learning rate and I used it for all of my other models.

Sequence length $l$ and batch size $n$ are two important hyperparameters as well. $n \times l$ is the number of images that are fed into the neural network during every mini-batches. It must not be too big because of the memory limitation of the GPU. However it can not be too small either otherwise the network will not learn enough information for each iteration to update itself. $n$ must be bigger than 1 as I mentioned above. $l$ is the number of state (sequential dimension) of Recurrent Neural network. I have tried different combination with three different $n$ values: 2,4,6 and three different $l$ values: 60,80,100. The difference in $l$ doesn't have a big impact on the result however $n$ with value 4 does outperform other choices in terms of converging speed and loss value at the end of training process.

By using early stop strategy, I found most of models will converge at 40 epochs, however, Resnet took longer time which is 60 epochs. So for all the models except Resnet, I choose 50 as my epochs number and for Resnet it's 60.

The size of the input image is not important in this project. The original VGGFACE\_GRU model used $96 \times 96$, however, the image I get after the Face Detection is $112 \times 112$. I have tried both of these sizes and it doesn't make much difference to the result. However training with $96 \times 96$ image is slightly faster than $112 \times 112$.  

\subsection{Training Process}
The code in the training scripts has been written by myself. I have used a configuration file to manage different hyperparameters and network model types. The script will run the training process for $n$ iterations where n is the number of epochs. During each epoch, the loss of each batch will be calculated out and recorded, then the loss will be back propagated to the network and the network will update itself. After one epoch has finished, it will also evaluate the network using both the training set and the test set.  I used the CCC to evaluate the network as I mentioned before. The better the network has performed on test data, the higher the score of the CCC (or small 1-CCC) that will be generated out. I have also work out the MSE for both the test and training sets, and the lower the MSE is, the better performance the model had. All of these results will be printed out as a checkpoint notes for me to plot after the training finished. 
Finally, the model will be saved after every 10 epochs with a unique file name so that I can load it whenever It's necessary.

Because I have been provided with two GPU to use, I usually ran the training script with two different configuration values at the same time. I planned to fix the network model type first then test with different hyperparameters. After I found the optimal combination of hyperparameters, I then used another model type. In the end, I would have the optimal results for all the network models I have mentioned before and I can compare them to give a conclusion about the best performing network on the dataset.

\subsection{Training result}
I evaluated the training result using the test and training CCC value. I will provide the plotted result with some explanations for different networks in the following subsections and I will conclude them with more specific data in the evaluation section.

\subsubsection{VGG16}
VGG16 is the network that Dimitrios has been using for the Aff-wild database. It has already been proven to be a good solution according to the result from his report. As my database contains continuous images from the videos, the CNN-RNN structure is expected to have a better performance than the CNN. But I have still trained the two different structures with their corresponding optimal hyperparameters in order to obtain a clear comparison.

\begin{figure}[h]
\centering
\begin{subfigure}{.5\textwidth}
  \centering
7  \includegraphics[width={0.9\textwidth},height={0.7\textwidth}]{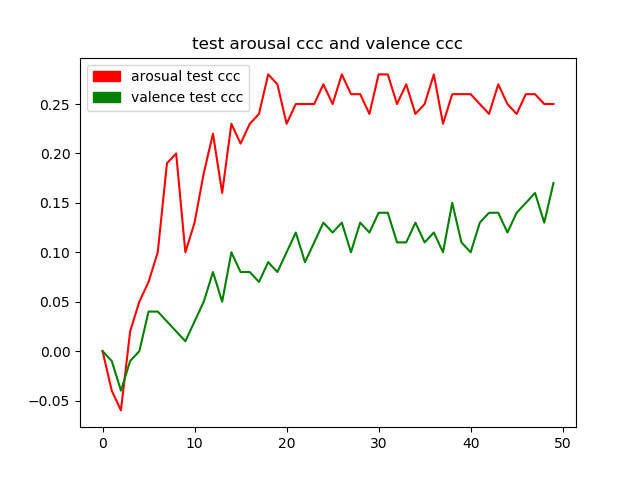}
  \caption{VGG16 arousal and valence CCC}
  \label{fig:sub1}
\end{subfigure}%
\begin{subfigure}{.5\textwidth}
  \centering
  \includegraphics[width={0.9\textwidth},height={0.7\textwidth}]{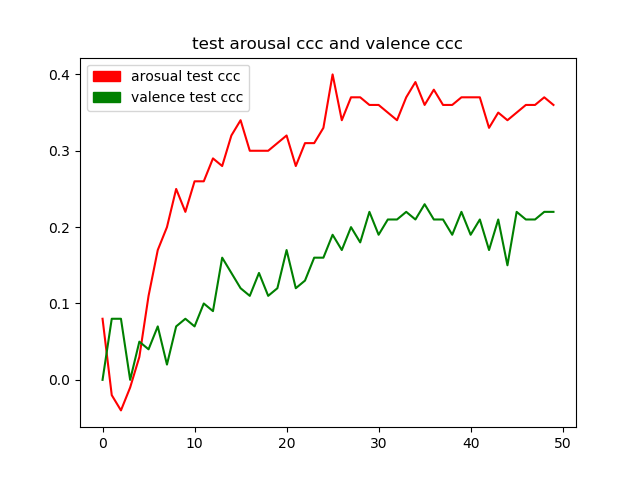}
  \caption{CNN-RNN arousal and valence CCC}
  \label{fig:sub2}
\end{subfigure}
\end{figure}

From the plot we can observe that the CNN-RNN model (VGG\_GRU) has a higher CCC in both valence value and arousal value compared to the CNN model. This result is expected because of the continuity of the data , this is the same as the other CNN models so I will only provide the result of CNN-RNN mode in other subsections. We can also observe that the arousal value has much higher CCC than valence value. This result is different from the result in Aff-wild database. According to Dimitrios, the Aff-wild database network has better performance in valence to arousal prediction. I have checked by predicting the emotion of the static image and both of the arousal and valence seems correct, so the values I am getting is right. Plus I am using exactly the same vgg\_gru network model as him, therefore, I can only consider this to be a unique property of my database. 

\begin{figure}[h]
\centering

  \includegraphics[width={0.6\textwidth},height={0.5\textwidth}]{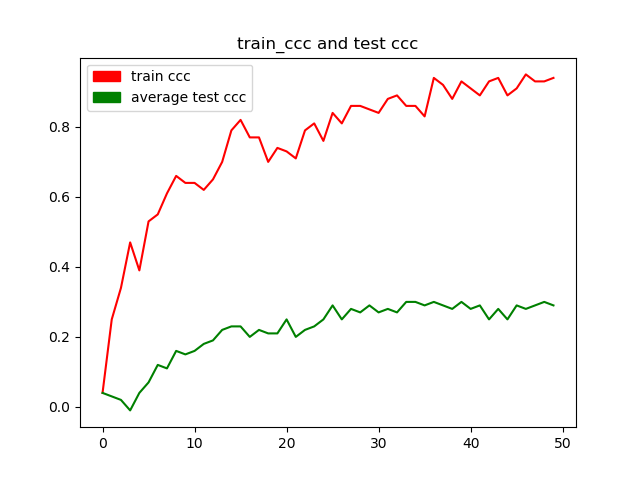}
  \caption{VGG16\_GRU training and testing CCC}

\end{figure}
The CCC in the above plot is the average CCC of valence CCC and arousal CCC. We can notice the average testing CCC reached 0.3 in the last few epochs (0.29 to be specific) while the training CCC reached almost 0.95. There is no obvious caused by overfitting so I can consider 0.3 is the highest test CCC I can achieve with my database. This result is almost the same as the VGG16\_GRU model using the Aff-wild database \cite{Aff-wild}  

\subsubsection{ResNet}
As I mentioned before, Resnet is designed for handling complicated tasks with deeper architecture. It's often expected to outperform other CNN model. Yet this is not the case in this project. Same as the VGG16\_GRU, I also used GRU as the RNN units to implement the CNN-RNN model.

\begin{figure}[h]
\centering
\begin{subfigure}{.5\textwidth}
  \centering
  \includegraphics[width={0.9\textwidth},height={0.7\textwidth}]{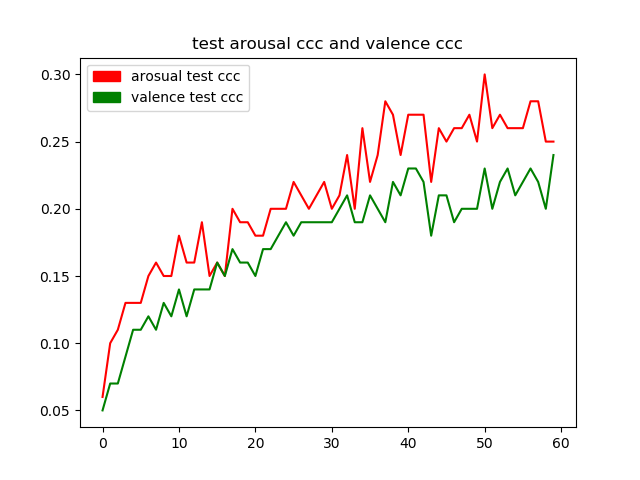}
  \caption{Resnet\_GRU test arousal and valence CC}
  \label{fig:sub1}
\end{subfigure}%
\begin{subfigure}{.5\textwidth}
  \centering
  \includegraphics[width={0.9\textwidth},height={0.7\textwidth}]{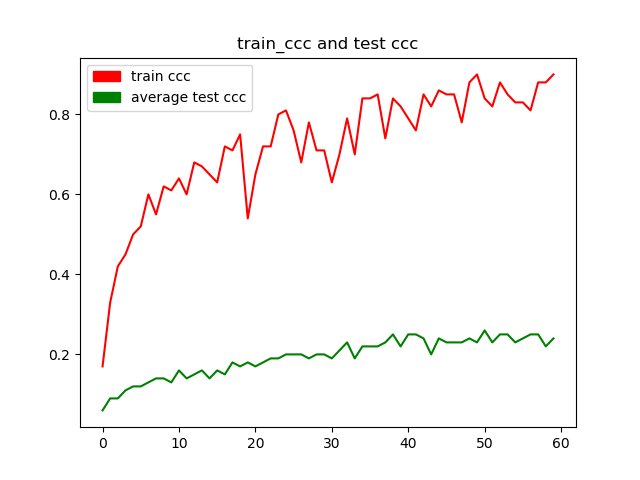}
  \caption{Resnet\_GRU testing and training CCC}
  \label{fig:sub2}
\end{subfigure}
\end{figure}

From the plot we can see the testing CCC only reached around 0.2 (0.24), again, this is almost the same result from Aff-wild Resnet\_GRU model. Until this point, I hadn't managed to get any higher results than it. However, from the left plot, we can observe the difference between Valence CCC and Arousal CCC is not as big as VGG16. This is a good sign, we can only predict the correct emotion if we can classify both valence and arousal correctly. At the end of this subject, both values should have a good CCC rather than a big difference between them. However, research needs to be done to discover how one can improve them at the same time.

\subsubsection{Alexnet}
Alexnet was not mentioned in the Aff-wild report, however after trying it with different hyperparameters and with some modifications to structure itself. The Alexnet\_GRU network did give me the best result compared to the other CNN-RNN architectures.  

\begin{figure}[h]
\centering
\begin{subfigure}{.5\textwidth}
  \centering
  \includegraphics[width={0.9\textwidth},height={0.7\textwidth}]{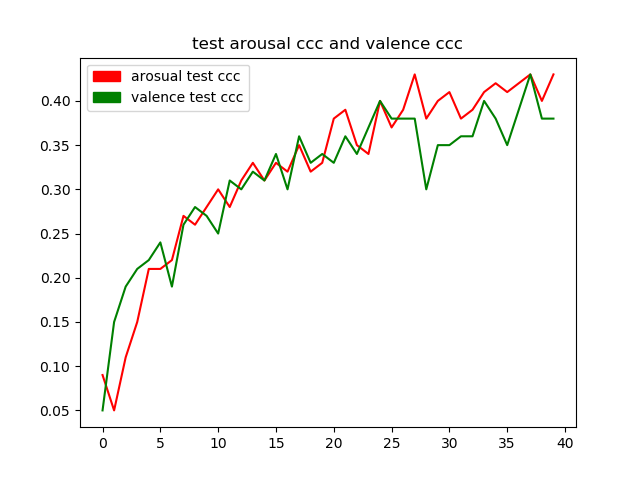}
  \caption{Alexnet\_GRU test arousal and valence CC}
  \label{fig:sub1}
\end{subfigure}%
\begin{subfigure}{.5\textwidth}
  \centering
  \includegraphics[width={0.9\textwidth},height={0.7\textwidth}]{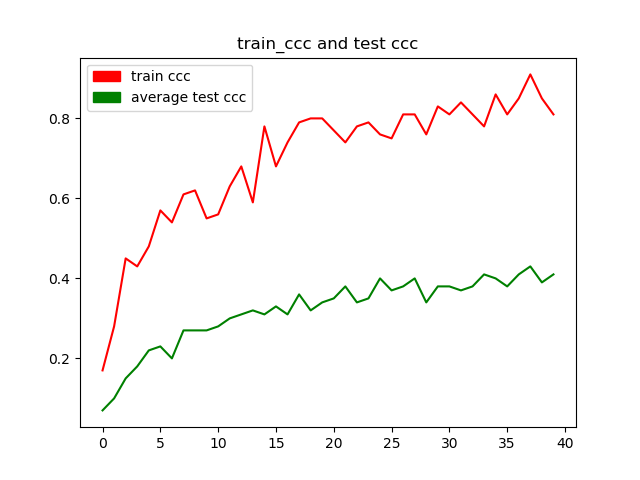}
  \caption{Alexnet\_GRU testing and training CCC}
  \label{fig:sub2}
\end{subfigure}
\end{figure}

The average test CCC achieved was 0.43 which is extremely high compared to other models. What's more interesting is that the valence CCC and arousal CCC are more similar to each other during the training process compared to other models. This is not expected because, as I mentioned in the Alexnet introduction , VGG16 outperformed Alexnet when both of them used Imagenet database. I assumed Alexnet was not as good as VGG16 in object classification but better in emotion classification. However, since there is no existing Alexnet\_GRU structure for me to use, I implemented this structure all by myself. What I did differently from VGG16\_GRU model is that I had two full connected layers after the series of CNN layers. I wondered if this could be the reason that Alexnet outperformed VGG. Unfortunately, I did test with VGG16\_GRU model with two full connected layers, and the result of this model is not even as good as the one full connected layer version. This is an interesting observation, the explanation is not possible to be given because of the constraints of my project.

\subsection{Training with pre-trained model}
Training a pretrained model with the new database is a method mentioned in the Aff-wild database report. I obtained the weights of  CNN layers from pretrained model, and use those weights as the initial weights of the CNN layers in my CNN-RNN model. Despite the catastrophic forgetting problem that neural network has , it still improved the performance by giving a good starting values to the weights since it's already a well-trained model. In my case, I used the VGG16 pretrained model that trained with Imagenet database to classify the objects. ImageNet is an image database organised according to the WordNet hierarchy (currently only the nouns), in which each node of the hierarchy is depicted by hundreds and thousands of images.\cite{imagenet} I have also used an Aff-wild pre-trained VGG16 model \cite{aff-wild11}. This model is provided by Dimitrios, and it's expected to outperform all the other models because it's a well-trained model that designed to generate the valence value and arousal values from images.

\subsubsection{ImageNet VGG16}
I firstly loaded the pre-trained weights into the VGG16\_GRU model that I have implemented. Then I trained the whole CNN-RNN architecture together with my database (You can also keep the CNN part untrainable and only train RNN layers).

\begin{figure}[h]
\centering
\begin{subfigure}{.5\textwidth}
  \centering
  \includegraphics[width={0.9\textwidth},height={0.7\textwidth}]{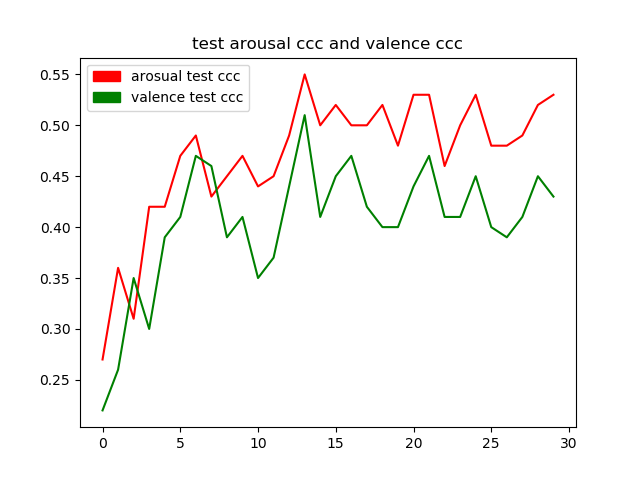}
  \caption{Pretrained VGG16 Imagenet Model test arousal and valence CC}
  \label{fig:sub1}
\end{subfigure}%
\begin{subfigure}{.5\textwidth}
  \centering
  \includegraphics[width={0.9\textwidth},height={0.7\textwidth}]{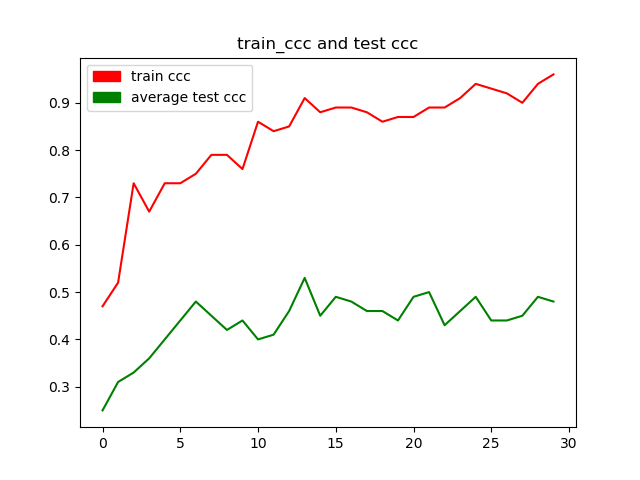}
  \caption{Pretrained VGG16 Imagenet Model testing and training CCC}
  \label{fig:sub2}
\end{subfigure}
\end{figure}

We can notice that the test CCC increases dramatically in the first 5 epochs then stay at 0.48 while the training CCC keep increasing. There is apparently a big improvement in performance and the difference between arousal and valence CCC is smaller than the result we get from VGG16 model. This is expected because firstly, the model is well-trained so it already has the ability to classify the object. Secondly, the model is trained with the database that has much more data than my database has, so the performance is indeed better than the model only trained with my database.

\subsubsection{Aff-wild VGG16 model}
This model is the VGG16 model Dimitrios used to train with Aff-wild database and according the report, it achieved 31 testing CCC in the end. \cite{multi-component} 

\begin{figure}[h]
\centering
\begin{subfigure}{.5\textwidth}
  \centering
  \includegraphics[width={0.9\textwidth},height={0.7\textwidth}]{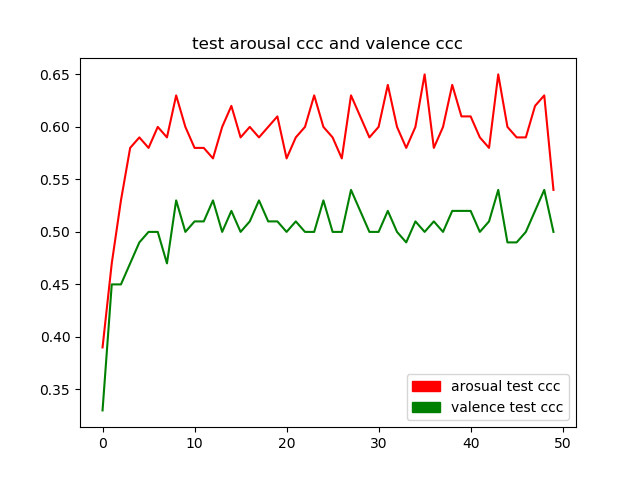}
  \caption{VGG16 Aff-wild Model test arousal and valence CC}
  \label{fig:sub1}
\end{subfigure}%
\begin{subfigure}{.5\textwidth}
  \centering
  \includegraphics[width={0.9\textwidth},height={0.7\textwidth}]{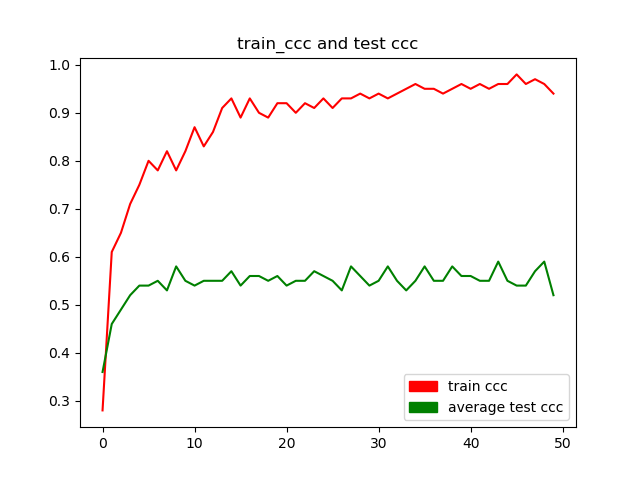}
  \caption{VGG16 Aff-wild Model testing and training CCC}
  \label{fig:sub2}
\end{subfigure}
\end{figure}

As expected, the performance of this model is better than the other models even in the first few epochs. This is because the pretrained model already has the ability to predict the accurate arousal and valence value from images. The pretrained model is a CNN architecture, so the performance is improved by adding RNN layers after it and training with more data from my own database. The average testing CCC achieved is 0.6 which is a really good result. However, it still has the problem with a big difference between the valence CCC and arousal CCC. 

\subsection{Evaluation}
I firstly evaluate the model I obtained with three values : Testing CCC values (The higher it is, the better performance it has), Testing MSE values (the lower it is, the better performance it has), and the Testing CCC that I tested the model with the Aff-wild database and Affectnet database. The Aff-wild database has more than 800000 images so the expected result should be much lower than the testing CCC, Affectnet uses statics image instead of continuous images from video. So I cannot test it with CNN-RNN model but CNN model.

Then I would like to show the predicted values for some of my own photos just to prove that this model can be used to classify the emotion from both videos and static images.
\newpage

\begin{table}[h]
\centering
\caption{The valence and arousal CCC}
\label{tab:my-table}
\begin{tabular}{@{}lll@{}}
\toprule
Model name                  & valence CCC & arousal CCC \\ \midrule
Trained from scratch \\\midrule
VGG16\_GRU                  & 0.22                    &     0.37                \\
ResNet\_GRU                 &  0.23                   &   0.25                  \\
AlexNet\_GRU                &  0.43                   &    0.42                 \\\midrule
Used pretrain model \\ \midrule
VGG16\_Imagenet\_pretrained &  0.44                   &   0.51                  \\ 
VGG16\_Affwild\_pretrained  &  0.53                   &     0.62                \\ \bottomrule
\end{tabular}
\end{table}

\begin{table}[h]
\centering
\caption{The valence and arousal MSE}
\label{tab:my-table}
\begin{tabular}{@{}lll@{}}
\toprule
Model name                  & valence MSE & arousal MSE \\ \midrule
Trained from scratch \\\midrule
VGG16\_GRU                  & 0.05                    &     0.04              \\
ResNet\_GRU                 &   0.06                 &   0.05                 \\
AlexNet\_GRU                &  0.05                  &   0.04                \\\midrule
Used pretrain model \\ \midrule
VGG16\_Imagenet\_pretrained &  0.05                   &   0.03                 \\ \midrule
VGG16\_Affwild\_pretrained  &  0.04                  &    0.02               \\ \bottomrule
\end{tabular}
\end{table}

By comparing the results from those models, one can notice that in the models that trained from scratch-which means they were only used to train my database. Alexnet\_GRU outperformed all the other models in terms of testing CCC values. However, the MSE values are similar between them. Since I used CCC as a loss function instead of MSE, it's reasonable that MSE is similar while CCC has a huge difference. Even though, we can make sure the difference between the result value and labelled value is not big because the MSE is small.

Comparing the result from models that used pretrained models. We noticed that those model performs better than the models trained from scratch. This is expected because as I mentioned ,the pretrained model is already a fine tuned model which has been trained with a much larger database. However VGG16 trained with Imagenet database has a slightly poorer performance than VGG16 trained with Aff-wild database. This is because the Aff-wild VGG model is designed to generate Valence and Arousal value while Imagenet VGG model is designed to classify the object instead. Aff-wild model is more suitable to our purpose.

I have tried to use some pre-trained Alexnet models. However the performance is nowhere near as good as the Alexnet model I trained from scratch. But I believe there could be some useful pre-trained Alexnet model suitable for this project and the performance could be better since Alexnet is the best model among the models trained from scratch.

\begin{table}[h]
\centering
\caption{The test CCC using Aff-wild database}
\label{tab:my-table}    
\begin{tabular}{@{}lll@{}}
\toprule
Model name                  & valence CCC & arousal CCC \\ \midrule
Trained from scratch \\\midrule
VGG16\_GRU                  & 0.14                    &    0.19             \\
ResNet\_GRU                 &   0.15                 &   0.16                 \\
AlexNet\_GRU                &  0.23                  &    0.28                \\\midrule
Used pretrain model \\ \midrule
VGG16\_Imagenet\_pretrained &  0.26                  &   0.27                 \\ 
VGG16\_Affwild\_pretrained  &  0.25                &    0.28               \\ \bottomrule
\end{tabular}
\end{table}

\begin{table}[h]
\centering
\caption{The test MSE using Aff-wild database}
\label{tab:my-table}
\begin{tabular}{@{}lll@{}}
\toprule
Model name                  & valence MSE & arousal MSE \\ \midrule
Trained from scratch \\\midrule
VGG16\_GRU                  & 0.12                    &    0.08            \\
ResNet\_GRU                 &   0.11                &   0.08                 \\
AlexNet\_GRU                &  0.11                   &   0.08                \\\midrule
Used pretrain model \\ \midrule
VGG16\_Imagenet\_pretrained &  0.10                  &   0.08                 \\ \midrule
VGG16\_Affwild\_pretrained  &  0.12                &    0.08               \\ \bottomrule
\end{tabular}
\end{table}

The CCC using the whole Aff-wild database as test data is smaller than the testing CCC I get from my own test set. There are few possible reasons for this difference. Firstly, the Aff-wild database has nearly 1000000 images while my test set only contains 130000 images, the  generalisation capabilities is limited since I only have 300000 images as a test set in my database. The second reason could be the difference in resolution. It is possible that the model trained with 4K resolution may have less accuracy when classify the images with lower resolution. The third reason is the difference in annotation between those two databases. The same emotion could have different valence and arousal values because of they have different annotators.  

The reason I want to evaluate my models on this database is because I want to be sure that the models work at a certain level with test images from another database (hence performing well in the general case). Even though the CCC values are lower than before, it's still a decent value according to the Aff-wild report. However, the AlexNet\_GRU along with the models that used pre-trained weights outperformed the VGG16\_GRU and ResNet\_GRU.

\begin{table}[h]
\centering
\caption{VGG16 CNN test CCC using my dataset and AffectNet database}
\label{tab:my-table}
\begin{tabular}{@{}lll@{}}
\toprule
Model name                  & valence CCC & arousal CCC \\ \midrule
VGG16 test on my database    & 0.14  &    0.26           \\
VGG16 test on AffectNet   &   0.11  &   0.20                \\
 \bottomrule
\end{tabular}
\end{table}

Firstly, the VGG16 CNN model is expected to have worse performance than VGG16\_GRU  model because of the continuity of the images from video can help a lot while using CNN-RNN model as I mentioned before. So the VGG only achieved 0.14 in valence CCC and 0.26 in arousal  CCC. VGG16 still have the problem with the difference in valence and arousal. Secondary. When I test the same model on AffectNet which is a completely new database, the result is not too bad compare the result from my own test set. Overall, the VGG16 CNN model also works with my database although it has worse performance than CNN-RNN model and it has been proved by testing on AffectNet.

\subsubsection{Classify my own emotion}
I selected the optimal model which used pretrained weights from VGG16 Aff-wild model to perform this evaluation. Since my model is a CNN-RNN structure and it's designed to classify the emotion of people in a continuous video. Thus,it's difficult to predict only one photo each time because the model require a fixed sequence length for RNN units to work. The way to solve it is to input a whole sequence of the same image into the model, let's say we have a sequence length of 80. Then the output will be an array with shape [80,2]. Where 2 refers to the pair of valence and arousal value. 

The first few pairs of values in the array is not reasonable because they don't have the hidden state in the beginning. However, in the last few pairs of values, the values tend to be stable and those are the final predictions for the emotion of the image. However, the result will not be as accurate as the ones I retrieved from test set, this is because the continuity of the images is one of the factors to improve the performance. However, the result is still good enough and I have provided some of my own photos along with the predicted values to make a comparison.

\begin{figure}[h]
\centering
\begin{subfigure}{.3\textwidth}
  \centering
  \includegraphics[width={0.9\textwidth},height={1.3\textwidth}]{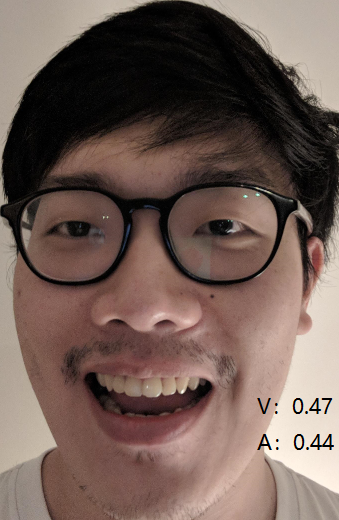}
  \caption{Happy}
  \label{fig:sub1}
\end{subfigure}%
\begin{subfigure}{.3\textwidth}
  \centering
  \includegraphics[width={0.9\textwidth},height={1.3\textwidth}]{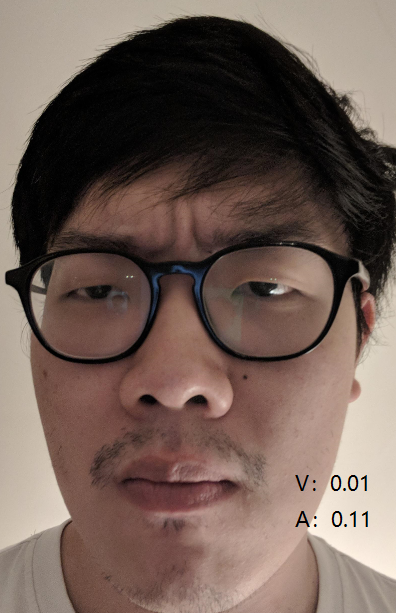}
  \caption{Worried}
  \label{fig:sub2}
\end{subfigure}
\begin{subfigure}{.3\textwidth}
  \centering
  \includegraphics[width={0.9\textwidth},height={1.3\textwidth}]{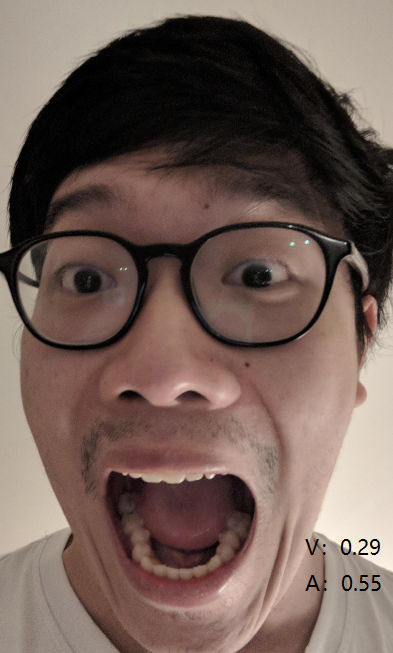}
  \caption{Surprised}
  \label{fig:sub2}
\end{subfigure}
\end{figure}

I noticed that the happy emotion has the highest valence value 0.47, and because I opened my mouth, the arousal is relatively high as well. However, when it comes to surprised, the arousal value increased dramatically to 0.55 while valence decreased to 0.29. For the worried photo. The valence is 0.01 while the arousal is 0.11. I expected the valence value to be negative, however this is the case as the performance is not perfect with a static image.

\begin{figure}[h]
\centering991
\begin{subfigure}{.3\textwidth}
  \centering
  \includegraphics[width={0.9\textwidth},height={1.3\textwidth}]{happy.PNG}
  \caption{Happy}
  \label{fig:sub1}
\end{subfigure}%
\begin{subfigure}{.3\textwidth}
  \centering
  \includegraphics[width={0.9\textwidth},height={1.3\textwidth}]{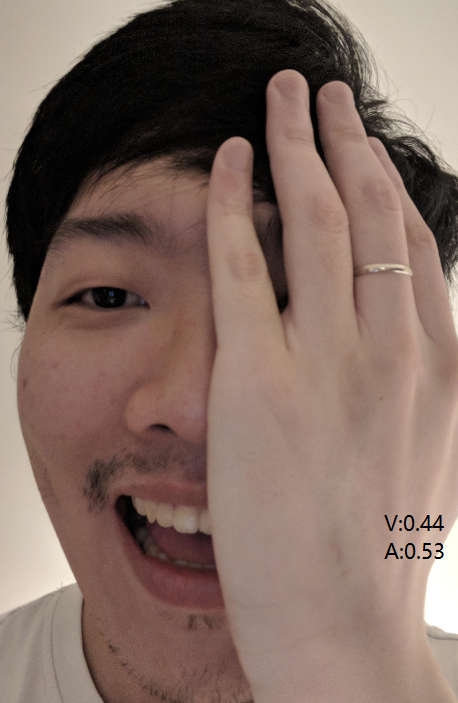}
  \caption{Half of my face}
  \label{fig:sub2}
\end{subfigure}
\begin{subfigure}{.3\textwidth}
  \centering
  \includegraphics[width={0.9\textwidth},height={1.3\textwidth}]{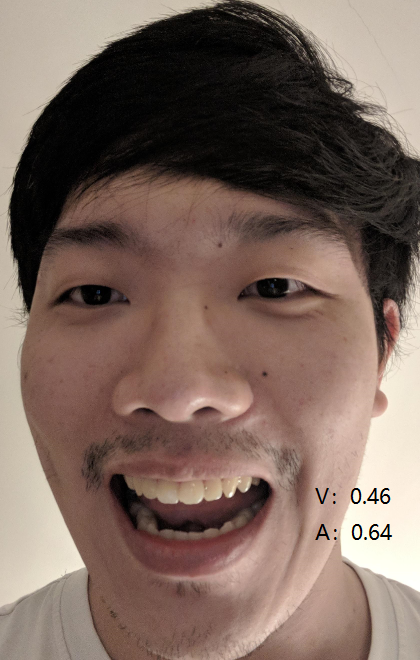}
  \caption{No glasses}
  \label{fig:sub2}
\end{subfigure}
\end{figure}

This group of photos show that the model can predict the emotion even when some part of the face is covered, or the accessory is removed. We notice the values didn't change a lot, however if we are predicting the values for a video, the values won't change at all while the person is covering their face.

\newpage
\section{Conclusion and Future  work}
This project is an extension of the Aff-wild challenge that has been done by Dimitrios and his team. I was aiming to verify the performance of different models that had been trained on the database with  in-the-wild images that have 4K resolution. Then find the optimal hyperparameters for each model. Finally, analysing the result and giving a reasonable explanation as to why these results were the cases. 

From the evaluation, I found several different models that perform well in this project, which means the CCC values between the predicted valence and arousal values and labelled values are relatively high.  And among those models that trained from scratch (without any pretrained weights loaded), Alexnet outperformed all the models that I have tried with.I also tried to load pretraiend weights into the model and train them use my database, the performance improved a lot if the pretrained model is suitable to this project. 

Overall, the result I have obtained from the experiment is good enough to conclude the models are able to predict out the accurate valence and arousal values given the images or videos. Those models are not the only accomplishment from this project, the database itself is valuable as well because it is the only database that contains in-the-wild images with 4k resolution. This database can be helpful if anyone else is interesting in this topic in the future once I uploaded them as open source file.

\subsection{Future work}
There are still some limitations in this project and they could be improved or fixed in the future with more resource and time. The first limitation is the size of the database. There are  only 403040 images in the database, which is not enough compare to Aff-wild database which has 996000 images in total. If I can extend the size into around 1000000 images, I can increase the generalisation capabilities of model and even improve the performance onto a new level.

The second limitation is that we can not use CCC to compare the performance of models that trained on different databases because that the different properties of databases could effect CCC. Which means I cannot  tell how much the 4k resolution helped in classification just by comparing the CCC between the result of my project and Aff-wild project. I should designed a better way to evaluate the improvement of 4K resolution in the beginning of this project. One possible approach could be : downscale the whole database into a smaller resolution, and treat it as the second database. Train a model with the second database, and then compare the CCC values with the same model that trained with the original database. However, I didn't left myself enough time on doing it in this project, but for future work, this can be considered as a way of evaluation.

Emotion recognition is always a big topic in computer vision and deep learning \cite{training,inter,on-line,adapt}. And this technique has already been used in the life.For example Some companies use this technique to classify the human emotion during a interview to tell the anxiety or the confidence level of interviewee. Some companies use it  to find out how how satisfied the people are during a meeting or activity. However, those images in the interview or the meeting are in the wild, and because of  the model I obtained from this project is trained with the in-the-wild database, I believe it can produce the result with higher accuracy in those scenarios. I hope this report and the models I have trained can give some information to the researchers who are working on this area, and improve the technique onto a higher level.

\newpage
\appendixpage
\appendix

\section{The links of videos I downloaded}
\begin{itemize}
\item https://www.youtube.com/watch?v=JwrYfolGwvw
\item https://www.youtube.com/watch?v=Q7NRAZ0o0jA 
\item https://www.youtube.com/watch?v=\_c4fRKJyI7Q 
\item https://www.youtube.com/watch?v=Y18l7XJiYRU 
\item https://www.youtube.com/watch?v=rIFbRlY3s1A
\item https://www.youtube.com/watch?v=UWQ3nfMii1s
\item https://www.youtube.com/watch?v=hhmjNH-VLZs
\item https://www.youtube.com/watch?v=uMVtitZoUcU 
\item https://www.youtube.com/watch?v=n8cQqjHgxKU
\item https://www.youtube.com/watch?v=a-lVNU9xoiU
\item https://www.youtube.com/watch?v=QkMFvx0UoUQ
\item https://www.youtube.com/watch?v=uS8evBfnFAg
\item https://www.youtube.com/watch?v=ZqgWan42AM4 
\item https://www.youtube.com/watch?v=Ie8Uv02pvZ4 
\item https://www.youtube.com/watch?v=Ssl\_tySn79E
\item https://www.youtube.com/watch?v=NxhyYcqsUpc
\item https://www.youtube.com/watch?v=G-Zdf1sxFhw 
\item https://www.youtube.com/watch?v=hWg6tc-MFDQ 
\item https://www.youtube.com/watch?v=UGmlHdRh5pE
\item https://www.youtube.com/watch?v=BSitS67ZQMs
\item https://www.youtube.com/watch?v=zYsVvTZLVwA
\item https://www.youtube.com/watch?v=ZDqI-8oGvH0
\item https://www.youtube.com/watch?v=6ia100kSzH8
\item https://www.youtube.com/watch?v=8V6yOw1kyC8
\item https://www.youtube.com/watch?v=GtLx5bSuFak 
\item https://www.youtube.com/watch?v=ZAHhfakGx\_M 
\item https://www.youtube.com/watch?v=IfExiziqN9c 
\item https://www.youtube.com/watch?v=ZtpOcMOxRM0 
\item https://www.youtube.com/watch?v=s5whE\_8HVX4
\item https://www.youtube.com/watch?v=q4gzDB4LGfs 
\item https://www.youtube.com/watch?v=UHN4hcvt9ag 
\item https://www.youtube.com/watch?v=JwZdOxbArYM 
\item https://www.youtube.com/watch?v=3hn6H5PUyI8 
\item https://www.youtube.com/watch?v=ULA6ebfJR0s 
\item https://www.youtube.com/watch?v=FxoQOvOBn8c
\item https://www.youtube.com/watch?v=VB3vBEJ99ck 
\item https://www.youtube.com/watch?v=vDlCTFOGUW0
\item https://www.youtube.com/watch?v=lXdoVSws5I4
\item https://www.youtube.com/watch?v=rWtIuzJAs-s
\item https://www.youtube.com/watch?v=TM\_xWR5fM7M 
\item https://www.youtube.com/watch?v=AuT6WuCYttk
\item https://www.youtube.com/watch?v=iafHJR2TBqE 

\item https://www.youtube.com/watch?v=6KcvJulvXK0 
\item https://www.youtube.com/watch?v=h1-0delJ3Gs
\item https://www.youtube.com/watch?v=N5RF4vy6a14 
\item https://www.youtube.com/watch?v=2eLFBpy-6o0 
\item https://www.youtube.com/watch?v=BiXP9WQaiII
\item https://www.youtube.com/watch?v=6Tzjeie8u8U
\item https://www.youtube.com/watch?v=bdnvkevz7os
\item https://www.youtube.com/watch?v=r9x-rhca-jY
\item https://www.youtube.com/watch?v=vL2YWDwOLmE
\item https://www.youtube.com/watch?v=zWGYenufq0I 
\item https://www.youtube.com/watch?v=h\_bktZoPwYk 
\item https://www.youtube.com/watch?v=SU5tYPXWxL4 
\item https://www.youtube.com/watch?v=MO3jtKUncDo 
\item https://www.youtube.com/watch?v=v3OYq4X9Hs8
\item https://www.youtube.com/watch?v=-0W7yjJ1hhE 
\item https://www.youtube.com/watch?v=k3MqxeRjMq0 
\item https://www.youtube.com/watch?v=bm2xcviOnJM
\item https://www.youtube.com/watch?v=P9HI-U4dwFs
\item https://www.youtube.com/watch?v=EPmZolscz2M
\item https://www.youtube.com/watch?v=0AgWi79b7zY 
\item https://www.youtube.com/watch?v=JtNpUMWvK5A
\item https://www.youtube.com/watch?v=2HcEa3kdmsw
\item https://www.youtube.com/watch?v=3KAX6NGyX-s
\item https://www.youtube.com/watch?v=4zqyr1m5xQQ
\item https://www.youtube.com/watch?v=dLWpngMUj\_M
\item https://www.youtube.com/watch?v=BCqo6swVOqQ
\item https://www.youtube.com/watch?v=5Yza0JBfxaI

\item https://www.youtube.com/watch?v=lj5yEb3oC-I
\item https://www.youtube.com/watch?v=6Jq9HdOyjEg 
\item https://www.youtube.com/watch?v=yrbMXKkj8fY 
\item https://www.youtube.com/watch?v=IGHK7t0KmlA
\item https://www.youtube.com/watch?v=hd\_F3YiM-gE 
\item https://www.youtube.com/watch?v=MAm8AZf1q2o 
\item https://www.youtube.com/watch?v=5QAObE2959I 
\item https://www.youtube.com/watch?v=ZiO4NB0saF0 
\item https://www.youtube.com/watch?v=ZsikYSSpAY0
\item https://www.youtube.com/watch?v=59vKzoo58qY 
\item https://www.youtube.com/watch?v=xKQZ0z86YBk
\item https://www.youtube.com/watch?v=rFClds\_bZHY 
\item https://www.youtube.com/watch?v=3sjETjMzNeI
\item https://www.youtube.com/watch?v=M0d88-FIVYc
\item https://www.youtube.com/watch?v=dpCpLK\_3974
\item https://www.youtube.com/watch?v=XcJusWTUDJI
\item https://www.youtube.com/watch?v=qCcYDcRdHqo 
\item https://www.youtube.com/watch?v=xH08VO534os
\item https://www.youtube.com/watch?v=EcE0EUa9hJw 
\item https://www.youtube.com/watch?v=p\_RC7mO8P5k 
\item https://www.youtube.com/watch?v=QOcP59bv9wQ
\item https://www.youtube.com/watch?v=w9a-fFCn7h8
\item https://www.youtube.com/watch?v=9rQxkK27eU4
\item https://www.youtube.com/watch?v=Zf3e\_9EmUo0 
\item https://www.youtube.com/watch?v=QsehRC61dMw
\item https://www.youtube.com/watch?v=JuPOEZnC0uo
\item https://www.youtube.com/watch?v=JuPOEZnC0uo
\item https://www.youtube.com/watch?v=\_GQgUi7UDrQ
\item https://www.youtube.com/watch?v=cenReC58Q7E
\item https://www.youtube.com/watch?v=VkZDBHunDT0
\item https://www.youtube.com/watch?v=\_hLVbd7xMpQ 
\item https://www.youtube.com/watch?v=o268qbb\_0BM 
\item https://www.youtube.com/watch?v=hL3HUp8TO9Q
\item https://www.youtube.com/watch?v=S2B8GBpF1Wg
\item https://www.youtube.com/watch?v=n\_y8b3PGb4Q
\item https://www.youtube.com/watch?v=sc4b9yvrB7c 
\item https://www.youtube.com/watch?v=mY0zxJVsu44
\item https://www.youtube.com/watch?v=P9PjqPVxc9E
\item https://www.youtube.com/watch?v=tzJ9okUO\_3s
\item https://www.youtube.com/watch?v=YQqlXf2mGrY 
\item https://www.youtube.com/watch?v=60JwgpO8zn8 
\item https://www.youtube.com/watch?v=6QSy06AYnZA
\item https://www.youtube.com/watch?v=YOemNXYdcMI 
\item https://www.youtube.com/watch?v=eCmPPz\_cf6A
\item https://www.youtube.com/watch?v=LYnSsvFaeF0
\item https://www.youtube.com/watch?v=xMMi\_iFsP\_I 
\item https://www.youtube.com/watch?v=w3dGCeFO2bg 
\item https://www.youtube.com/watch?v=OV1mVKLoCGI\&t=5s
\item https://www.youtube.com/watch?v=ev0AztYOU94 
\item https://www.youtube.com/watch?v=BESePFgt8nI
\item https://www.youtube.com/watch?v=lKEHRA1MOiU
\item https://www.youtube.com/watch?v=sgxx3rkS3rs
\end{itemize}
\end{document}